\pdfoutput=1
\documentclass[10pt]{article}
\pdfpageattr {/Group << /S /Transparency /I true /CS /DeviceRGB>>}

\usepackage{rotating}
\usepackage{graphicx}
\usepackage{subfigure}
\usepackage{amsmath}
\usepackage{amssymb}
\usepackage{multirow}
\usepackage{multicol}
\usepackage{algorithmic}
\usepackage{algorithm}
\usepackage{longtable} 
\usepackage{siunitx}
\usepackage{makecell}
\usepackage[top=3cm, bottom=3cm, left=3.5cm, right=3.5cm]{geometry}

\usepackage{url}

\def\RealSet{\mbox{I\hspace*{-0.16em}R}}

\def\path{\leadsto}

\def\RealSet{\mbox{$I\!\!R$}}

\def\shape#1{\rm }

\def\B0{\mbox{\boldmath $0$}}

\def\CalC{{\cal C}}

\def\CalF{{\cal F}}

\def\CalH{{\cal H}}
\def\CalI{{\cal I}}

\def\CalM{{\cal M}}
\def\CalN{{\cal N}}

\def\CalS{{\cal S}}
\def\CalT{{\cal T}}

\def\CalV{{\cal V}}

{\begin{list}{}{\setlength{\itemindent}{-\leftmargin}\setlength{\itemsep}{0pt}\setlength{\parsep}{0pt}}}%
{\end{list}}

\def\citet{\cite}
\def\citep{\cite}
\def\RealSet{\mbox{$I\! \! R$}}

\DeclareMathOperator*{\argmin}{argmin}

\begin{document}
%
\title{\huge\textbf{Learning to see like children: \\
proof of concept}}

\author{\small Marco Gori, Marco Lippi, Marco Maggini, Stefano Melacci\\
\small  Department of Information Engineering and Mathematics\\ 
\small University of Siena, Italy\\
\small  \texttt{\{marco,lippi,maggini,mela\}@diism.unisi.it}}
\date{}

%
%

\markboth{Technical Report}
{Shell \MakeLowercase{\textit{et al.}}: TITLE HERE!}
%


\maketitle


\begin{abstract}
In the last few years we have seen a growing interest in machine learning approaches to computer vision
and, especially, to semantic labeling.
Nowadays state of the art systems use deep learning on millions of labeled images with 
very successful results on benchmarks, though it is unlikely to expect similar results  in unrestricted visual environments.
Most of those sophisticated and challenging learning schemes extract regularities and perform 
symbolic prediction by essentially ignoring the inherent sequential structure of video streams. 
This might be a very critical issue, since any visual recognition process is remarkably 
more difficult when shuffling the video frames, which seems to be unnatural and hard even for humans.
Hence, it seems that most of current state of the art approaches to semantic labeling have been attacking 
a problem which is significantly harder than the one faced by humans.
Based on this remark, in this paper we propose a re-foundation of the communication protocol 
between visual agents and the environment, which is referred to as {\em learning to see like children}. 
Like for human interaction,  visual concepts are expected to be acquired by the agents solely by processing their own 
visual stream along with human supervisions  on selected pixels, instead of relying on huge labeled databases. 
We give a proof of concept that remarkable semantic labeling can emerge within this protocol by using only a few supervised
examples. This is made possible by fully exploiting the principle that in a learning environment 
based on a video stream, any intelligent agent willing to attach semantic labels to a moving pixel is 
expected to take coherent decisions with respect to its motion. Basically, the constraint of motion coherent 
labeling (MCL) virtually offers tons of supervisions, that are essentially ignored in most machine 
learning approaches working on big labeled data. MCL and other visual constraints, including those associated with 
object supervisions, are properly used with the context of the  theory of {\em learning from constraints}, that is
properly extended in the framework of lifelong learning, so as our visual agents are expected to {\em live in
their own visual environment} without distinguishing learning and test set.  The learning process
takes place in deep architectures under a developmental scheme that drives the progressive development
of visual representations in the layers of the net.\\
In order to test the performance of our {\em Developmental Visual Agents} (DVAs), 
in addition to classic benchmarking analysis, we open the doors of our lab, thereby allowing people 
to  evaluate DVAs by crowd-sourcing. No matter how efficient and effective our current DVAs are, 
the proposed communication protocol, as well as the
crowd-sourcing based assessment mechanism herein proposed might result in a paradigm shift in 
methodologies and algorithms for computer vision, since other labs might  
conceive  truly novel solutions within the proposed framework 
to face the long-term challenge of dealing with unrestricted visual environments. 
$\ $
\end{abstract}
\newpage
\section{Introduction}
\label{sec:intro}

Semantic labeling of pixels is amongst most challenging problems that are nowadays faced in computer vision. The availability of an enormous amount of image labels enables the application of sophisticated learning and reasoning models that have been proving their effectiveness in related applicative fields of AI. 

\marginpar{{\em Shuffling\\ frames makes\\ vision hard }}
Interestingly, so far, the semantic labeling of pixels of a given video stream has been mostly carried 
out at frame level. This seems to be the natural outcome of well-established pattern recognition methods
working on images, which have given rise to nowadays emphasis on collecting big labelled image 
databases~(e.g. \cite{imagenet}) with the purpose of devising and testing challenging machine learning algorithms.
While this framework is the one in which most of nowadays state of the art object recognition approaches 
have been developing, we argue that there are strong arguments to start exploring the more natural 
visual interaction that humans experiment in their own environment. To better grasp this issue, 
one might figure out what human life could have been in a world of visual information with shuffled frames. 
Any cognitive process aimed at extracting symbolic information from  images that are not frames of 
a temporally coherent visual stream would have been extremely harder than 
in our visual experience. Clearly, this comes from the information-based principle that 
in any world of shuffled frames, a video requires order of magnitude more information for its storing 
than the corresponding temporally coherent visual stream. 
As a consequence, any recognition process is remarkably 
more difficult when shuffling frames, and it seems that most of current state of the
art approaches have been attacking a problem which is harder than the one faced by humans. This
leads us to believe that the time has come for an in-depth re-thinking of machine learning 
for semantic labeling. As it will be shown in Section~\ref{sec:human}, we need a re-foundation of 
computational principles of learning under the framework of a human-like natural communication 
protocol to naturally deal with unrestricted video streams. 

\marginpar{{\em Beyond the\\ ``peaceful interlude"}}
From a rough analysis of the growing role played in the last few years by machine learning 
in computer vision, we can see that there is a rich collection of machine learning algorithms that have been 
successfully integrated into state of the art computer vision architectures.
On the other side, when the focus is on machine learning, vision tasks
are often regarded as yet other benchmarks to provide motivation for the proposed theory. 
However, both these approaches seem to be the  outcome of the bias coming 
from two related, yet different scientific communities. In so doing we are likely  missing 
an in-depth understanding of fundamental computational aspects of vision.
In this paper, we start facing the challenge of disclosing the computational basis of vision 
by regarding it as a truly learning field that needs to be attacked by an appropriate {\em vision learning theory}. 
Interestingly, while the emphasis on a general theory of vision was already the main objective 
at the dawn of the discipline~\cite{marr1982vision}, it has evolved without 
a systematic exploration of foundations in machine learning. 
When the target is moved to unrestricted visual environments and the emphasis is
shifted from huge labelled databases to a human-like protocol of interaction, {\em we need to go beyond 
the current peaceful interlude that we are experimenting in vision and machine learning}. 
A fundamental question a good theory is expected to answer is {\em why children can 
learn to recognize objects and actions from a few supervised examples, whereas nowadays
machine learning approaches strive to achieve this task}. In particular, why are 
they so thirsty for supervised examples? Interestingly, this fundamental difference  seems to be 
deeply rooted in the different communication protocol at the basis of the acquisition of visual
skills in children and machines. In this paper we propose a re-foundation of the communication protocol 
between visual agents and the environment, which is referred to as {\em learning to see like children} (L2SLC). 
Like for human interaction,  visual concepts are expected to be acquired by the agents solely by processing their own 
visual stream along with human supervisions  on selected pixels, instead of relying on huge labelled databases. 
In this new learning environment based on a video stream, any intelligent agent willing to attach 
semantic labels to a moving pixel is expected to take coherent decisions with respect to its
motion. Basically, any label attached to a moving pixel has to be the same during its motion\footnote{Interestingly, early studies on tracking exploited the invariance of brightness to 
estimate the optical flow~\cite{hornshunk}.}. Hence,  video streams provide a huge amount
of information  just coming from imposing  coherent labeling, which is likely to be the essential
information associated with visual perception experienced by any animal. Roughly speaking, 
once a pixel has been labeled, the constraint of coherent labeling virtually offers tons of other
supervisions, that are essentially ignored in most machine learning approaches working on 
big databases of labeled images. It turns out that most of the visual information to perform semantic labeling
comes from the motion coherence constraint, which explains the reason why children 
learn to recognize objects from a few supervised examples. The linguistic
process of attaching symbols to objects takes place at a later stage of children development, 
when he has already developed strong pattern regularities. We conjecture that, regardless of
biology, the enforcement of motion coherence constraint is a high level computational principle that plays 
the fundamental role for discovering pattern regularities. On top of the representation 
gained by motion coherence, the mapping to linguistic descriptions is dramatically simplified 
with respect to  machine learning approaches to semantic labeling based on huge
labeled image databases. This also suggests that the enormous literature on tracking is a mine of
precious results for devising successful methods for semantic labeling. 

\marginpar{{\em Deep learning\\ from visual\\ constraints}} 
The work described in this paper is rooted on the theory of \textit{learning from constraints} \cite{learningfromconstraints} 
that allows us to model the interaction of intelligent agents with the environment by means of constraints 
on the tasks to be learned. It gives foundations and algorithms  to discover 
tasks that are consistent with the given constraints and minimize a parsimony index. 
The notion of constraint is very well-suited to express both visual and linguistic granules of knowledge.
In the simplest case, a visual constraint is just a way of expressing the supervision on a labelled pixel, but 
the same formalism is used to express motion coherence, as well as 
complex dependencies on real-valued functions, that also include abstract logic formalisms\footnote{This is made possible by adopting the T-norm mechanism to express predicates by real-valued functions.} 
(e.g. First-Order-Logic (FOL))~\cite{diligenti2012bridging}. In addition to learning the tasks, like for kernel machines, given a new constraint, 
one can check whether it is compatible with the given collection of constraints~\cite{gori2013constraint}.  
While the representation of visual knowledge by logic formalisms is not covered in this paper, 
we can adopt the same mathematical 
and algorithmic setting used for representing the visual constraints herein discussed.
The main reason for the adoption of visual constraints is that they nicely address the chicken-and-egg 
dilemma connected with the classic problem of segmentation. 
The task of performing multi-tag prediction for each pixel of the input video stream,  
with semantics that involves different neighbors, poses strong restrictions on the computational mechanisms, thus 
sharing intriguing connections with biology. We use deep architectures that progressively learn convolutional filters 
by enforcing information-theoretic constraints, thus maximizing the mutual information between 
the input receptive fields and the output codes (Minimal Entropy Encoding, MEE \cite{mee}). 
The filters are designed in such a way that they are invariant under geometric (affine) transformations. 
The learned features lead to a pixel-wise deep representation that, in this paper, is used to predict semantic tags
by enforcing constraints coming from supervised pairs,  spatial relationships, and motion coherence. 
We show that the exploitation of motion coherence is the key to reduce the computational burden 
of the invariant feature extraction process. In addition, we enforce motion coherence within
the manifold regularization framework to express the consistency of high-level tag-predictions. 

\marginpar{{\em Life-long \\ learning}}
The studies on learning from constraints covered in~\cite{learningfromconstraints} lead to learning algorithms whose fundamental 
mechanism consists of  checking the constraints on big data sets\footnote{This holds true for soft-constraints, which are those of interest in this paper.}, which suggests that there are classic statistical learning principles behind the theory. 
It is worth mentioning that this framework of learning suggests to dismiss the difference between 
supervised and unsupervised examples, since the case of supervised pairs
is just an instance of the general notion of constraints. While this is an ideal view to embrace different visual
constraints in the same mathematical and algorithmic framework, clearly we need the re-formulation of the
theory to respond to the inherent on-line L2SLC communication protocol. Basically, the visual
agent is expected to collect continuously its own visual stream and acquire human supervisions on 
labels to be attached to selected pixels. This calls for lifelong learning computational schemes
in which the system adapts gradually to the incoming visual stream. In this paper  clustering 
mechanisms are proposed to store a set of templates under the restrictions imposed by the available memory budget. 
This allows us to have a stable representation to handle transformation invariances and perform real-time predictions 
while the learning is still taking place. It turns out that, in addition to dismissing the difference between supervised and
unsupervised examples, the lifelong computational scheme associated with our visual agents leads to
dismissing also the difference between learning and test set. These intelligent agents undergo developmental
stages that very much resemble humans'~\cite{gori:semantic-based} and, for this reason, throughout the paper, they are
referred to as  \textit{Developmental Visual Agents} (DVAs). 

\marginpar{{\em Proof of \\ concept}} 
This paper provides a proof of concept of the
feasibility of learning to see like children with  a few supervised examples of visual concepts. 
In addition to the CamVid benchmark, which allows us to relate DVAs performance to the
literature, we propose exploring a different experimental validation that seems to resonate perfectly
with the proposed L2SLC communication protocol. Given any visual world, along with a related
collection of object labels, we activate a DVA which starts living in its own visual environment by
experimenting the L2SLC interaction. Interestingly, just like children, as time goes by, the DVA is
expected to perform object recognition itself.  Now, it takes a little to recognize if a child is blind 
or visually impaired. The same holds for any visual agent, whose skills can be quickly evaluated.
Humans can easily and promptly judge the visual skills by checking a visual agent at work. 
The idea can fully be grasped\footnote{From the site, you can also download a library for 
testing DVAs in your lab.} at
\begin{center}
\url{http://dva.diism.unisi.it} \ 
\end{center}
where we open our lab to people willing to evaluate DVAs performance. Interestingly, the same
principle can be used for any visual agent which experiments the L2SLC protocol. 
The identity of people involved in the assessment is properly verified so as to avoid  unreliable 
results. A massive crowd-sourcing could give rise to a truly novel performance evaluation
that could nicely complement benchmark-based assessment. DVAs are only expected to be the first
concrete case of living visual agents that lean under the L2SLC that are evaluated by crowd-sourcing.
Other labs, either by using their current methods and technologies or by conceiving novel solutions, might 
be stimulated to come up with their own solutions in this framework, which could lead to 
further significant improvements with respect to those reported in this paper. 
Results on the CamVid benchmark confirm the soundness of the approach, especially when considering the simple and uniform mechanism to acquire
visual skills.
\subsection{Related work}
There are a number of related papers  with our approach.
Notably, the idea of receptive field can be traced back to the studies of Hubel and Wiesel \cite{hubel1962receptive} 
and, later on, it was applied to computer vision in the Neocognitron model \cite{fukushima1988neocognitron}.
Convolutional neural networks \cite{lecun1998gradient} have widely embraced this idea, and they recently leaded to state-of-the art results in object recognition on the ImageNet data \cite{krizhevsky2012imagenet,imagenet}.  Those results were also extended toward tasks of localization and detection~\cite{overfeat}. Recently, some attempts of transferring the internal representation to other tasks were studied in~\cite{laptevcvpr2014} with interesting results.
Other approaches develop hierarchies of convolutional features without any supervision. The reconstruction error and sparse coding are often exploited \cite{topographicdictionary,deepconvolutionalsparsecoding}, as well as solutions based on 
K-Means~\cite{coates2011analysis}.
Different  autoencoders have been proposed in the last few years~\cite{stackedautoencoders,contractiveautoencoders}, 
which have been tested on very large scale settings~\cite{catsfacesICML}. The issue of representation 
has been nicely presented in~\cite{bengio2013representation}, which also contains a comprehensive review of these approaches.
Some preliminary results concerning low-level features developed by DVAs have been presented in~\cite{ourECCVpaper,ourICIAPpaper}. The notion of invariance in feature extraction has been the subject of many analyses 
on biologically inspired models~\cite{serre2005object,poggio2009invariance}.
Invariances to geometric transformations are strongly exploited in hand-designed low level features, 
such as SIFT~\cite{sift}, SURF~\cite{surf}, and HOG~\cite{hog}, as well as in the definition of similarity 
functions~\cite{tangentdistance}.
We share analogies with the principles inspiring scene parsing approaches, which aim at assigning a tag to each pixel in an image. Recent works have shown successful results on classical benchmarks~\cite{liu2011nonparametric,tighe2013superparsing,tighe2013finding}, 
although they seem to be very expensive in terms of computational resources. 
Fully supervised convolutional architectures were exploited for this task in~\cite{farabet2013learning}, while a successful approach based on random forest is given is~\cite{PelilloTPAMI2014}.

The theory of \textit{learning from constraints}~\cite{learningfromconstraints} was applied 
in several contexts and with different types of knowledge, such as First-Order Logic 
clauses~\cite{diligenti2012bridging,gori2013constraint} and visual relationships in object recognition~\cite{melacciicann}. 
In the case of manifold regularization based constraints \cite{melacci2011laplacian}, our on-line learning system was evaluated using heterogenous data, showing promising results~\cite{frandinaICANN}.
Finally, the notion of constraint is used in this paper to model motion coherence, 
thus resembling what is usually done in optical flow algorithms \cite{hornshunk,opticalflowGPU}. 
Motion estimation in DVAs is part of the feature extraction process; some qualitative results can be found 
in~\cite{ourCVPRpaper}.


%
%
%

\section{En plein air}
\label{sec:human}
The impressive growth of computer vision systems has strongly benefited from the massive diffusion of
benchmarks which, by and large, are regarded as fundamental tools for performance evaluation.  
However, in spite of their apparent indisputable dominant role in the understanding of progress in computer
vision, some criticisms have been recently raised~(see e.g.~\cite{benchmarks}), which suggest that the time has come to
open the mind towards new approaches. 
%
%
\begin{figure}
\centering
\includegraphics[width=0.75\textwidth]{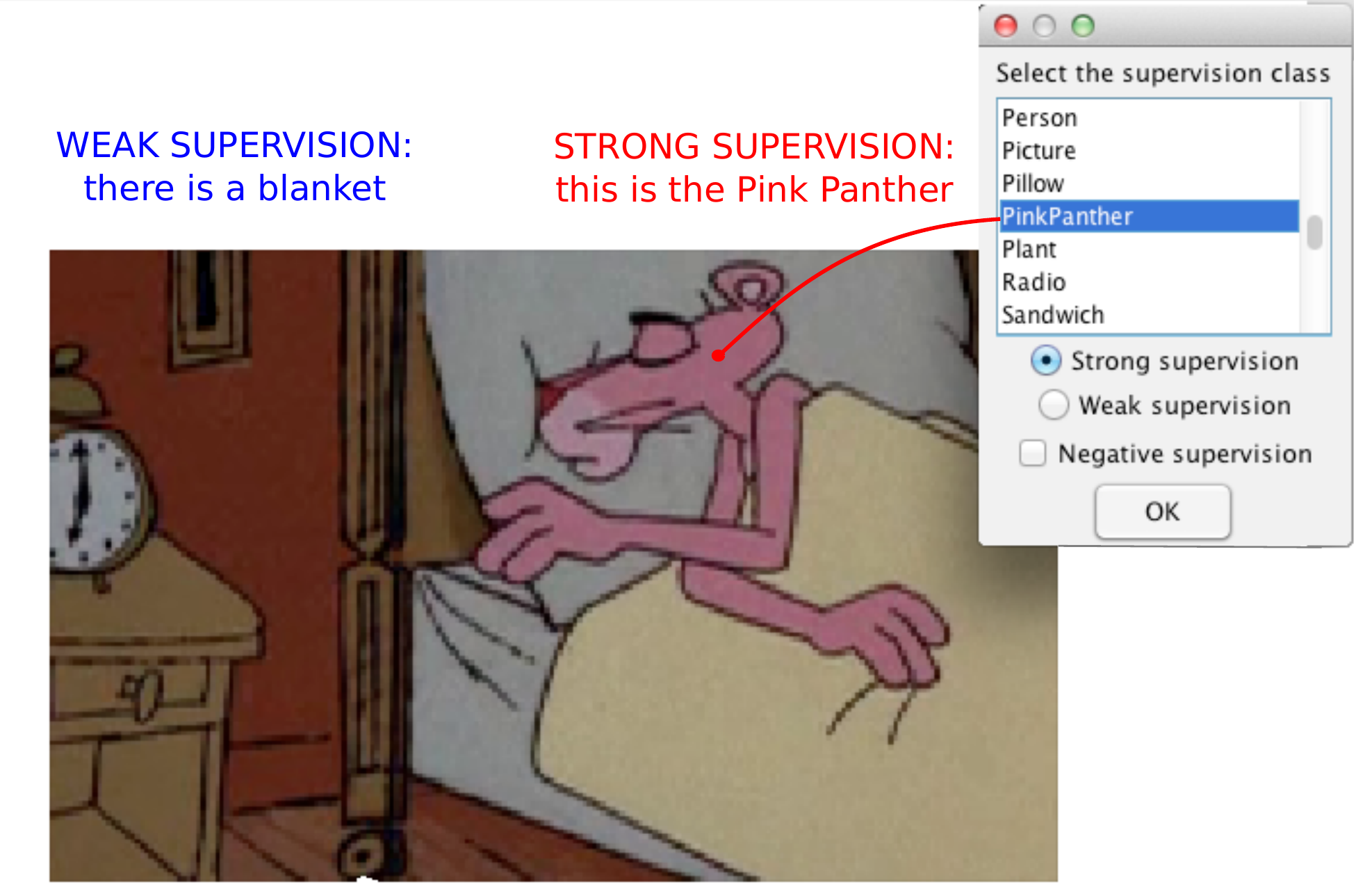}
\caption{{\em En plein air} in computer vision according to the current 
interpretation given in this paper for DVAs at~\texttt{http://dva.diism.unisi.it}.
Humans can provide strong and weak supervisions 
(Pink Panther cartoon, \textcopyright\ Metro Goldwyn Mayer).\label{fig:supervision}}
\end{figure}
%
%
The benchmark-oriented attitude, which nowadays dominates the computer vision community, 
bears some resemblance to the influential testing movement in psychology which has its roots in the 
turn-of-the-century work of Alfred Binet on IQ tests. In both cases, in fact, we recognize a familiar pattern: 
a scientific or professional community, in an attempt to provide a rigorous way of assessing the 
performance or the aptitude of a (biological or artificial) system, agrees on a set of standardized 
tests which, from that moment onward, becomes the ultimate criterion for validity.
As well known, though, the IQ testing movement has been severely criticized by many a scholar, 
not only for the social and ethical implications arising from the idea of ranking human beings on a 
numerical scale but also, more technically, on the grounds that, irrespective of the care with which 
these tests are designed, they are inherently unable to capture the multifaceted nature of real-world phenomena. 
As David McClelland put it in a seminal paper which set the stage for the modern competency movement in the U.S.,  
Òthe criteria for establishing the ÔvalidityÕ of these new measures really ought to be not grades in school, 
but Ôgrades in lifeÕ in the broadest theoretical and practical sense.Ó  
Motivated by analogous concerns, we maintain that the time is ripe for the computer vision 
community to adopt a similar Ògrade-in-lifeÓ attitude towards the evaluation of its systems and algorithms. 
We do not of course intend to diminish the importance of benchmarks, as they are indeed invaluable 
tools to make the field devise better and better solutions, but we propose we should use them 
in much the same way as we use school exams for assessing the abilities of our children: once they pass 
the final one, and are therefore supposed to have acquired the basic skills, we allow them to find a job in the real world.
Accordingly, in this paper we  open the doors of our lab to go {\em en plein air}, thereby allowing people all over the world 
to freely play and interact with the visual agents that will grow up in our lab\footnote{The idea of {\em en plein air}, along with the underlined relationships with human intelligence
has mostly come from enjoyable and profitable discussions with Marcello Pelillo, who also coined the term
and contributed to the above comment during our common preparation of a Google Research Program Award
proposal. In the last couple of years, the idea has circulated during the GIRPR meetings thanks also to 
contributions of Paolo Frasconi~\url{http://girpr.tk/sites/girpr.tk/files/GIRPRNewsletter_Vol4Num2.pdf} 
and Fabio Roli~\url{https://dl.dropboxusercontent.com/u/57540122/GirprNewsletter_V6_N1.pdf.}.}. 
%
%
\marginpar{{\em Evaluation by\\ crowd-sourcing}}
A crowd-sourcing performance evaluation scheme can be conceived where registered people can 
inspect and assess the visual skills of software agents. A prototype of a similar evaluation scheme
is proposed in this paper and can be experimented at~\url{http://dva.diism.unisi.it}.
The web site hosts a software package with a graphical interface which can be used to interact 
with the DVAs by providing supervisions and observing the resulting predictions. 
The human interaction takes place at symbolic level, where semantic tags are attached to 
visual patterns within a given frame. In our framework, users can provide two kinds of supervisions: 
\begin{description}
\item{$i$} {\em Strong supervision} -  one or more labels are attached to a specific pixel of a certain frame to express 
the presence of an object at different levels of abstraction; 
\item{$ii$} {\em Weak supervision} -  one or more labels are attached to a certain frame to 
express the \textit{presence} of an object, regardless of its specific location in the frame.
\end{description} 
The difference between strong and weak supervision can promptly be seen in Figure~\ref{fig:supervision}.
Strong supervision conveys a richer message to the agent since, in addition to the object labels,
 also specifies the location. In the extreme case, it can be a pixel, but labels can also be attached
 to areas aggregated by the agent. Weak supervision has a higher degree of abstraction, 
 since it also requires the agent to locate object positions. 
 In both cases, an object is regarded as a structure identified by a  position
 where one can attach different labels which depend on the chosen context.  
 For example, in Figure~\ref{fig:supervision}, the labels {\em eye} and {\em Pink Panther} 
 could be attached during strong supervision while pointing to a Pink Pather's eye. 
 Weak supervision can easily be provided by a microphone while wearing a camera,
 but it is likely to be more effective after strong supervisions have already been provided,
 thus reinforcing visual concepts in their initial stages.
 A visual agent is also expected to \textit{ask} to take the initiative by
 asking for supervision, and it is also asked to carry out an active learning scheme.
 The results reported in this paper for DVAs are only based on strong supervision,  but 
 the extension to weak supervision is already under investigation
 \footnote{An interesting solution for constructing visual environments for 
 massive experimentation is that of using computer graphics tools. In so doing,
 one can create visual world along with symbolic labels, that are available 
 at the time of visual construction. Clearly, because of the abundance of 
 supervised pixels, similar visual environment are ideal
 for statistical assessment. This idea was suggested independently by
 Yoshua Bengio and Oswald Lanz.}.

While this paper gives the proof of concept of the L2SLC protocol along with the en plein air crowd-sourcing 
assessment, other labs could start exposing their models and technologies, as well as new solutions,
within the same framework.

\section{Architectural issues}
\label{sec:architecture}
%
\begin{figure*}
\includegraphics[width=1.0\textwidth]{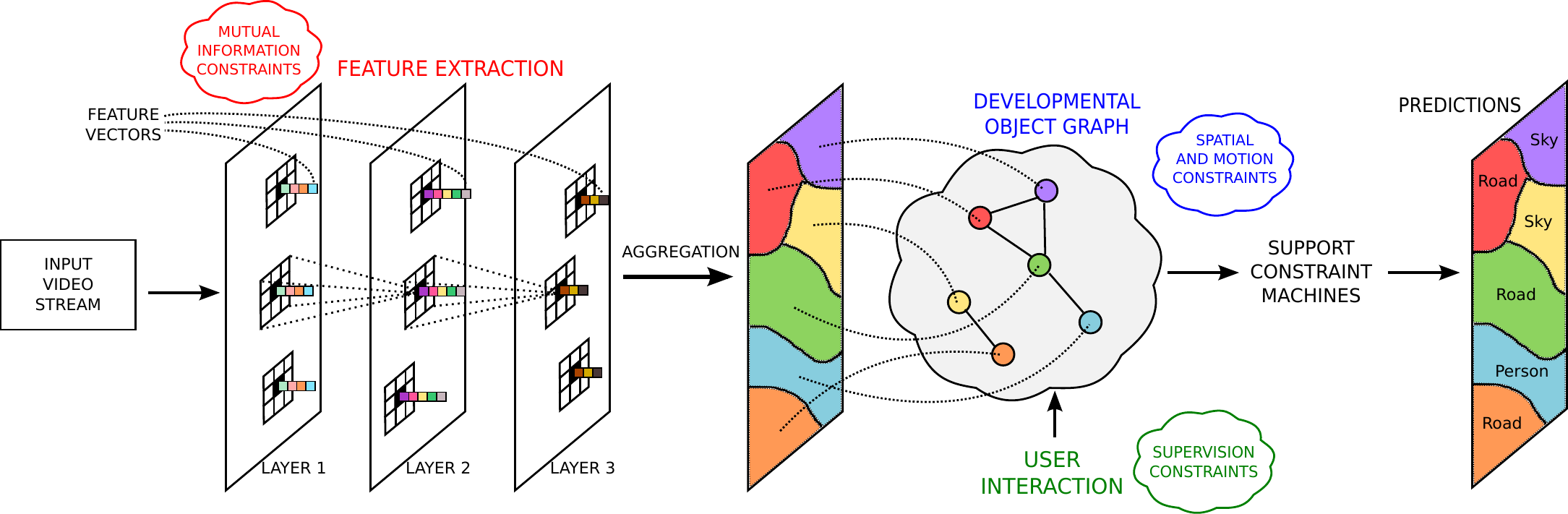}
\caption{An overview of the architecture of a Developmental Visual Agent. A deep network
learns a smooth satisfaction of the visual constraints within the general framework proposed in~\cite{learningfromconstraints}. An appropriate interpretation of the theory is given to allow the implementation of a truly lifelong learning scheme. 
\label{fig:overall}}
\end{figure*}
The architecture of the whole system is depicted in Figure~\ref{fig:overall}. Basically, it is a deep network whose layers contain features that are extracted from receptive fields. As we move towards the output, the hierarchical structure makes it possible to virtually cover larger and larger areas of the frames. 
\marginpar{{\em Pixel-based \\ features}}
Let $\CalV$ be a video stream, and $\CalV_t$ the frame processed at time $t$. For each layer $\ell = 1, \ldots, L$, a DVA extracts a set of $d^\ell$ features $\overline{f}^\ell_j(x,\CalI_t^{\ell})$, $j = 1, \ldots, d_\ell$, for each pixel $x$, where $\CalI^\ell_t$ is the input of layer $\ell$ at time $t$, i.e. $\CalI_t^1 = \CalV_t$. The features are computed over a neighborhood of $x$ (receptive field) at the different levels of the hierarchy. To this aim, we model a \textit{receptive field} of $x$ by a set of $\CalN$ Gaussians $g_k$, $k=1,\ldots,\CalN$, located nearby the pixel.  We define {\em receptive input} of $x$, denoted by  $\xi(x,\CalI_{t}^{\ell}) = \left[ \xi_{1}(x,\CalI_{t}^{\ell}),\ldots,\xi_{\CalN}(x,\CalI_{t}^{\ell}) \right]^{\prime}$, the value
\begin{equation}
\xi_{k}(x, \CalI_{t}^{\ell}) = g_k \otimes \CalI_{t}^{\ell} \propto \int_{\CalI_{t}^{\ell}} e^{-\frac{\| x_{k} + x - y \|^2}{2 \eta^2}} \CalI_{t}^{\ell}(y) dy \ .
\label{ri1}
\end{equation}
The receptive input $\xi(x,\CalI_{t}^{\ell})$ is a filtered representation of the neighborhood of $x$, which expresses a degree of image detail that clearly depends on the number of Gaussians $\CalN$ and on their variance $\eta$. 
Notice that, $\forall j=1,\ldots,d$,
the association of each pixel $x$ to its receptive input $\xi$ 
induces the function $f_j^{\ell}(\xi(x,\CalI_t^{\ell})):=\overline{f}_j^{\ell}(x,\CalI_t^{\ell})$.  
Although the position of the centers $x_k$ is arbitrary, we select them on a uniform grid of unitary edge centered on $x$.
From the set of features learned at layer $\ell$, $f^\ell_1(x,\CalI_{t}^{\ell}), \ldots, f^{\ell}_{d_\ell}(x,\CalI_{t}^{\ell})$, a corresponding set of probabilities is computed by the softmax function 
\begin{equation}
	p^{\ell}_j = \frac{e^{f^{\ell}_j}}{\sum_{i=1}^{d^{\ell}} e^{f^{\ell}_i}}
\label{SoftMaxEq}
\end{equation}
so that all the $d^\ell$ features satisfy the probabilistic normalization, thus competing
during their development. The feature learning process takes place according to
information-theoretic principles as described in Section~\ref{sec:features}.
In order to compact the information represented by the features, we project them onto a space of lower dimensionality by applying stochastic iterations of the NIPALS (non-linear iterative partial least squares) algorithm~\cite{nipals} to roughly compute the principal components over a time window.
Moreover, in order to enhance the expressiveness of DVA features, they are partitioned into $C^\ell$ subsets (\textit{categories}), so as the learning process takes place
in each category $c$  by producing  the probability vector $p_c^\ell(x,\CalI_t)$ of $d^{\ell}_c$ elements,
independently of the ones of other categories, with $\sum_{c=1}^{C^\ell} d^{\ell}_c = d^\ell$.
Different categories are characterized by the different portions of the input taken into
account for their computation. For example, at the first layer, each category can operate on a different input channel (e.g., for an RGB encoding of the input video stream) or on different projections of the input. 

\marginpar{{\em Region-based \\ features}}  
After features have been extracted at pixel-level, an aggregation process takes place 
for partitioning the input frame into ``homogeneous regions". To this aim, we extend the graph-based region-growing algorithm by Feszenwalb and Huttenlocher~\cite{hanselgretel} in order to  enforce motion coherence in the development. The original algorithm in~\cite{hanselgretel} starts with each pixel belonging to a distinct region, and then progressively aggregates pixels by evaluating a dissimilarity function based on color similarity (basically, Euclidean distance between RGB triplets or grayscale values). We enrich this measure by decreasing (increasing) the dissimilarity score of pixels whose motion estimation is (is not) coherent. The idea is to enforce the similarity of neighbor pixels locally moving to the same direction. The similarity is also increased for those pairs of neighbor pixels that at the previous frame were assigned to the same static region (no motion). 
Once the regions have been located, properly region-based features are constructed which summarizes in different ways the associated information. 

\marginpar{{\em Developmental \\ Object Graph}} 
The regions correspond to visual patterns that are described in terms of 
an appropriate set of features and that are stored into the nodes $v_j \in V$ 
of a graph, referred to as \textit{Developmental Object Graph} (DOG). The
edges of the graph represent node similarity as well as
motion flow information between two consecutive frames (see Section~\ref{sec:symbolic}). 

The symbolic layer  is associated with the functions 
$f_1(v_j),\ldots,f_{\omega}(v_j)$ that are defined on the DOG nodes. These functions are also forced to respect constraints 
based on the spatio-temporal manifold induced by the DOG structure. 
We overload the symbol $f$ to define both pixel-wise low-level features 
and high-level tag predictors to refer to functions that are developed
under learning from constraints. 
We assume that  $f$ operates in a {\em transductive environment} both 
on the receptive inputs and on the DOG nodes. As it will be shown later,
this allows us to buffer predictions and perform real-time response.

\section{Learning from visual constraints}
\label{sec:lfc}
The features and the symbolic functions involved in the architectural structure of
Fig.~\ref{fig:overall} are learned within the framework of learning from
constraints \cite{learningfromconstraints}. 
In particular, the feature functions $f_j^{\ell}(\xi)$  turn out to be the smooth maximization of the mutual information between the output codes and the input video data (Section \ref{sec:features}).
The high-level symbolic functions $f_z(v)$ are subject to constraints on motion and spatial coherence, as well as to constraints expressing supervised object (Section \ref{sec:symbolic}). Additional
visual constraints can express relationships on the symbolic functions, including logic
expressions. The constraints enrich their expressiveness with the progressive exposition to
the video so as to follow a truly lifelong learning paradigm. 
DVAs are expected to react and make predictions at any time, 
while the learning still evolves asynchronously.
\marginpar{{\em Parsimonious \\ constraint \\ satisfaction }} 
Let $f$ be a vectorial function $f=[f_1,\ldots,f_n]$ such that  $f \in \CalF$.
We introduce its degree of parsimony by means of an appropriate norm\footnote{See~\cite{Gori-NC-2013} for an in-depth discussion on the norms to express 
parsimony in Sobolev spaces and for the connections with kernel machines.} 
$\| f \|$ on $\CalF$. We consider a collection of
visual constraints $\CalC$ indexed by $q =1,\ldots,m$
and indicate by $\mu_{\CalC}^{(q)}$ a penalty that
expresses their degree of  fulfillment. The problem of learning from (soft) constraints
consists of finding
\begin{equation}
f^{*} = \arg \min_{f \in \CalF} \left\{ \| f \|^{2} + \sum_{q=1}^{m} \mu_{\CalC}^{(q)}(f) \right\} \ .
\label{eq:problem}
\end{equation}
Its general treatment is given in~\cite{learningfromconstraints}, where a 
functional representation is given along with algorithmic solutions. In this paper
we follow one of the proposed algorithmic approaches that is based on considering
the sampling of constraints.  
While $f_j(\xi(x,\CalI_t))$ and $f_z(v)$ operate on different domains they can both
be determined by solving eq.~\ref{eq:problem} and, therefore, for the sake of simplicity,
we consider  a generic domain  $X = \{x_1,\ldots,x_N \}$, without making any distinctions
between feature-based and high-level symbolic functions. 
In addition, the hypothesis of sampling the constraints makes it possible to
reduce the above parsimony index to the one induced by a Reproducing Kernel Hilbert Space (RKHS) 
$\CalH$, that is $\|f\|^2 = \lambda \sum_{k=1}^{n} \| f_k \|_{\CalH}^2$, where $\lambda$ is 
the classic regularization parameter. 
In \cite{learningfromconstraints} it is shown that, regardless of the kind of constraints, 
the solution of~\ref{eq:problem}  is given in terms of a Support Constraint Machine (SCM).

\marginpar{{\em Transductive \\ environments}} 
A representer theorem is given that extends the classical  kernel-based representation
of  traditional learning from examples. 
In particular, $f_k^{*}$ can be given an optimal representation that is
based on the following finite expansion
\begin{equation}
f_{k}^{*} = \sum_{i=1}^{N} \zeta_{ik} k(x_i, \cdot) \ ,
\label{eq:solution}
\end{equation}
where $k(\cdot,\cdot)$ is the kernel associated with the selected norm\footnote{Under certain boundary conditions, in~\cite{Gori-NC-2013}, 
it proven that $k(\cdot,\cdot)$ is the Green function
of the differential operator $L$, where $L=P^{\star} P$, $P^{\star}$ is the adjoint of $P$, 
and $ \| f \|^{2}:=\langle P f,P f\rangle$.}, and $\zeta_{ik}$ 
are the parameters to be optimized. 
They can be obtained by gradient-based optimization of the 
function that arises when plugging~\ref{eq:solution} into~\ref{eq:problem}, so as
the functional $ \| f \|^{2} + \sum_{q=1}^{m}\mu_{\CalC}^{(q)}(f)$, collapses to finite dimensions. 

%
A very important design choice of DVAs is that they operate into a transductive
environment. This is made possible by clustering the incoming data
into the set of representative elements $X$.
\marginpar{{\em On-line \\ learning }} 
Clearly, the clustering imposes memory restrictions, and it turns out to be important
to define a budget to store the elements of $X$, as well as their removal policy. 
The clustering differs in the case of $f_j^{\ell}(\xi)$ and $f_z(v)$, and it
will be described in Section \ref{sec:features} and \ref{sec:symbolic}, respectively.
The values $f(x)$ are cached over $x \in X$ after each update of 
$\zeta_{ik}$, so that DVAs make predictions at any time, independently of
the status of the optimization process. 
The on-line learning consists of updating $\zeta_{ik}$ along with the data stream. 
The parameters $\zeta_{ik}$ associated with newly introduced representatives are set to zero, 
to avoid abrupt changes of $f$.





\section{Learning invariant features}
\label{sec:features}

\begin{figure}
\centering
\includegraphics[width=0.75\textwidth]{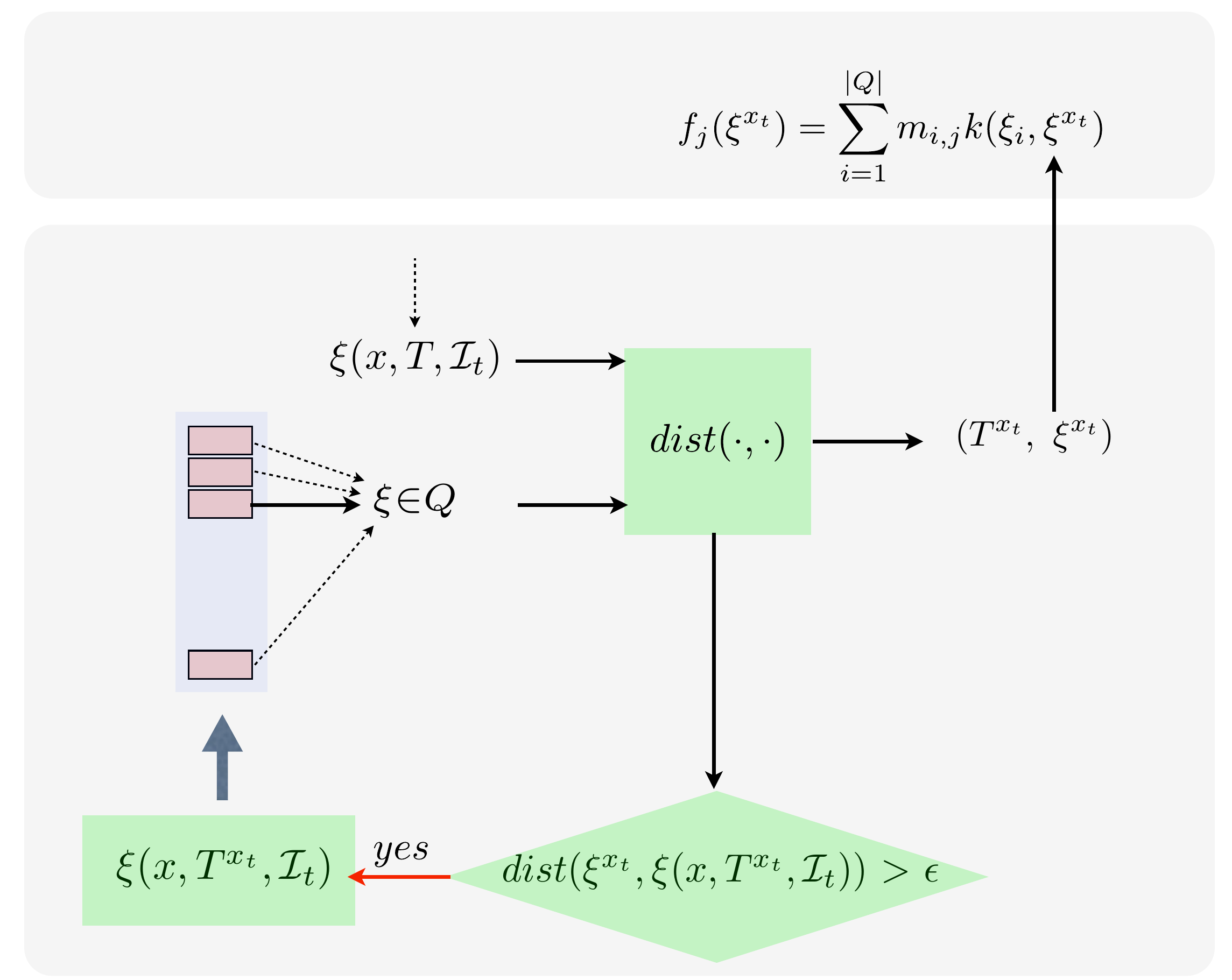} 
\vskip -3mm
\caption{The two-step process for the computation of pixel-based features.
First, invariance is gained by searching in $Q$ for the optimal receptive input
$\xi^{x_{t}}$. Second, the parameters $m_{i,j}$ are learned in the framework of the 
support constraint machines, with mutual information constraints.
Notice that an efficient search in $Q \times \CalT$ (see dotted lines) is made possible by a local search based on motion coherence.}
\label{fig:PBF-architecture}
\end{figure}

\indent In this section we describe the on-line learning algorithm used by DVAs for 
developing the pixel-level features $f_j^{\ell}(\xi)$. As sketched in Figure~\ref{fig:PBF-architecture},
the features are learned by means of a two stage process. First, DVAs gain invariance 
by determining an appropriate receptive input and, then, they learn the local pattern shapes
by a support constraint machine.  

Let us start with the stage devoted to discovering invariant receptive inputs. 
Given a generic layer and category\footnote{For the sake of simplicity, in the rest of this section, 
we drop the layer and category indices.}, for each pixel $x$, we want to incorporate
the affine transformations of the receptive field into the receptive input $\xi(x,\CalI_t)$.  
Since any 2D affine map $A$  can be rewritten as the composition of three 2D transformations and a scale parameter, then we can express $A = \sigma R_{\varphi_1} U_{\varphi_2}  R_{\varphi_3}$, where $\sigma>0$ and $R_{\varphi_1}$, $U_{\varphi_2}$, $R_{\varphi_3}$ are 
\begin{eqnarray}
\nonumber R_{\varphi_1} &=& \left[ \begin{array}{cc} \cos \varphi_1& -\sin \varphi_1 \\ \sin \varphi_1 & \cos \varphi_1 \end{array} \right],\ U_{\varphi_2} = \left[ \begin{array}{cc} \frac{1}{\cos\varphi_2}& 0 \\ 0 & 1 \end{array} \right],\\
\nonumber R_{\varphi_3} &=& \left[ \begin{array}{cc} \cos \varphi_3& -\sin \varphi_3 \\ \sin \varphi_3 & \cos \varphi_3 \end{array} \right],
\end{eqnarray}
with $\varphi_1\in[0,2\pi]$, $\varphi_2\in[0,\frac{\pi}{2})$, and $\varphi_3\in[0,\pi)$ \cite{asift,Melacci-WP1-2014}. These continuous intervals are discretized into grids $\Phi_1, \Phi_2, \Phi_3$, and, similarly, we collect in $\Sigma$ a set of discrete samples of $\sigma$ (starting from $\sigma=1$). The domain $\CalT = \Sigma \times \Phi_1 \times \Phi_2 \times \Phi_3$ collects all the possible \textit{transformation tuples}, where $\varphi_1$, $\varphi_2$, $\varphi_3$ and $\sigma$ can be considered as hidden variables for $\xi$, depending on pixel $x$. Given a tuple $T\in \CalT$, we can calculate each component of the receptive input $\xi(x, T, \CalI_t)$ as
\begin{equation}
\xi_{k}(x, T, \CalI_t) \propto \int_{\CalI_t} e^{-\frac{\| \sigma R_{\varphi_1} U_{\varphi_2}  R_{\varphi_3} x_{k} + x - y \|^2}{2 \sigma^2\eta^2}} \CalI_t(y) dy \ ,
\label{ri2}
\end{equation}
where the value of $\sigma$ affects both the width of the Gaussians and their centers, and the dependency of $\varphi_1$, $\varphi_2$, $\varphi_3$ and $\sigma$ from $x$ has been omitted to keep the notation simpler. Note that computing the receptive input for all the pixels $x$ and for all the transformations in $\CalT$ only requires to perform $|\Sigma|$ Gaussian convolutions per-pixel, independently of the number of centers $\CalN$ and on the size of the grids $\Phi_1, \Phi_2, \Phi_3$, since only $\sigma$ affects the shape of the Gaussian functions\footnote{The non-uniform scaling of $U_{\varphi_2}$ should generate anisotropic Gaussians (see \cite{asift}), that we do not consider here both for simplicity and to reduce the computational burden.}.
The receptive input can also include invariance to local changes in brightness and contrast, that we model by normalizing $\xi(x,T,\CalI_t)$ to zero-mean and unitary $L_2$ norm\footnote{Those receptive inputs that are almost constant are not normalized. The last feature of each category, i.e., $f_{d^{\ell}}(\xi)$, is excluded from the learning procedure and hard-coded to react to constant patterns.}. 

For any pair $(t,x)$, the tuple $T^{x_t}$ is selected in order to minimize the mismatch of $\xi(x,T^{x_t},\CalI_t)$ from a discrete sampling of the receptive inputs processed up to the current frame-pixel. Let $Q$ be such a collection of receptive inputs\footnote{We do not explicitly indicate the dependance of $Q$ on the frame and pixel indices to keep the notation simpler.}, 
and let $dist(\cdot,\cdot)$ be a metric on $Q$.
\marginpar{{\em Dealing with \\ invariance}} 
Formally, we associate $T^{x_t}$ to $x \in \CalI_t$ such that
\begin{equation}
\left( T^{x_t},\ \xi^{x_t} \right) = \argmin_{\ T \in \CalT,\ \xi\in Q} \ dist(\xi, \xi(x,T,\CalI_t)) \ ,
\label{eq:fullmatch}
\end{equation}
being $\xi^{x_t} \in Q$ the closest element to $\xi(x,T^{x_t},\CalI_t)$. Such matching criterion allows us to associate each pixel to its nearest neighbor in $Q$ and also to store the information of its transformation parameters $T^{x_t}$. We introduce a tolerance $\epsilon$ which avoids storing near-duplicate receptive inputs in $Q$.
Clearly, the choice of $\epsilon$  determines the sampling resolution, thus defining the clustering process. 
After having solved eq. (\ref{eq:fullmatch}), if $dist(\xi^{x_t}, \xi(x,T^{x_t},\CalI_t)) > \epsilon$, then $\xi(x,T^{x_t},\CalI_t)$ is added to $Q$, otherwise it is associated with the retrieved $\xi^{x_t}$ (see Figure~\ref{fig:PBF-architecture}). 
The data in $Q$ are distributed on a $(\CalN-2)$-sphere  of radius $1$, 
because of the $L_2$ normalization and the mean subtraction. When $dist(\cdot, \cdot)$ is chosen as the Euclidean distance, a similarity measure based on the inner product $ \langle \cdot,\cdot \rangle$ can be equivalently employed to compare receptive inputs, such that the constraint $dist(\xi_i,\xi_j) > \epsilon$ can be verified as $\langle\xi_i, \xi_j\rangle < \gamma_{\epsilon}$.
The set $Q$ is an $\epsilon$-net of the subspace of $\RealSet^{\CalN}$ that contains all the observed receptive inputs. Such nets are standard tools in metric spaces, and they are frequently exploited in searching problems because of their properties~\cite{ourCVPRpaper}. For instance, it can be easily shown that there exists a \textit{finite} set $Q$ for any processed video stream.

\marginpar{{\em The mutual \\ information \\ constraint }} 
The second stage of learning consists of discovering a support constraint machine which 
operates on the set $Q$ within the framework of Section \ref{sec:lfc}. The idea is that of
maximizing the mutual information (MI) of  $Y(f) \in \RealSet^{d}$, which represents 
the codebook of features for the considered category, and $X$, which 
represents data stored in $Q$. When searching for smooth solutions $f$, this problem is
an instance of Minimal Entropy Encoding (MEE)~\cite{mee} and, more generally, of
learning with Support Constraints Machines~\cite{learningfromconstraints}. 
Let us denote by $H(\cdot)$ and $H(\cdot| \cdot)$ the entropy and the conditional entropy
so as  
\[
	MI(Y(f);X)=H(Y(f)) - H(Y(f) | X).
\]
The constraint $MI(Y(f);X) = \log d$ enforces the maximization of the mutual information, and we can use eq. (\ref{eq:problem}) to define the penalty $\mu_{\CalC}^{(1)}(f) := \mu_{MI}(f)$ where
\begin{eqnarray*}
\mu_{MI}(f) &=& | H(Y(f)) - H(Y(f) | X) -  \log d | \\
&=& | MI(Y(f);X) - \log d | \\
&=& -MI(Y(f);X)\ .
\end{eqnarray*}
where the solution is given by eq. (\ref{eq:solution}), i.e. by the kernel expansion 
\[
	f_j (\xi) = \sum_{i=1}^{|Q|} m_{ij} k({\xi}_i, \xi),
\]
with $\xi_i \in Q$.

Finding a solution to eq. (\ref{eq:fullmatch}) for all pixels in a given frame can be speeded up by a pivot-based mechanism \cite{ourCVPRpaper}, but it still quickly becomes computationally intractable, when the resolution of the video and the cardinality of the set of transformation tuples $\CalT$ achieve reasonable values. 

\marginpar{{\em Motion \\ estimation: \\  tractability \\ of invariance }} 
In order to face tractability issues, we exploit the inherent coherence of video sequences so as the pairs $(T^{x_{t-1}},\xi^{x_{t-1}})$ are used to compute the new pairs $(T^{x_{t}},\xi^{x_{t}})$. The key idea is that the scene smoothly changes in subsequent frames and, therefore, at a certain pixel location $x$, we are expected to detect a receptive input which is very similar to one of those detected in a neighborhood of $x$ in the previous frame. In particular, we impose the constraint that both the transformation tuple $T^{x_{t}}$ and the receptive input $\xi^{x_{t}}$ should be (almost) preserved along small motion directions.
Therefore,  we use a heuristic technique which performs quick local searches that can provide good
approximations of the problem stated by~eq. (\ref{eq:fullmatch}),
while greatly significantly speeding up  the computation\footnote{We refer to~\cite{ourCVPRpaper} for the details.}. It is worth mentioning that the proposed heuristics to determine invariant parameters also yields, as a byproduct, motion estimation for all the pixels of any given frame. Strict time requirements in real-time settings can also be met by partitioning the search space into mini-batches, and by accepting sub-optimal solutions of the nearest neighbor computations within a pre-defined time budget.

\marginpar{{\em Blurring}} 
At the beginning of the life of any visual agent, $Q$ is empty, and new samples are progressively 
inserted as the time goes by. From the dynamic mechanism of feature development, 
we can promptly realize that the clustering process, along with the creation of $Q$, turns out to be strongly 
based on the very early stage of life. In principle, this does not seem to be an appropriate 
developmental mechanism, since the receptive inputs that become cluster representatives
in $Q$ might not naturally represent visual patterns that only come out later in the agent life.
In order to face this problem we propose using a \textit{blurring scheme} such that $Q$  ends up
into a nearly stable configuration only after a certain {\em visual developmental time}. We initially set the variance scaler $\eta$ of the Gaussian filters of eq. (\ref{ri2}) to a large value, and progressively decrease it with an exponential decay that depends on $t$. This mechanism produces initial frames (layer inputs) strongly blurred, so that only a few $\xi$'s are added to $Q$ for each frame (even just one or none\footnote{The tuple assigned to the first addition to $Q$ is arbitrary.}). As $\eta$ is decreased, the number of items in $Q$ grows until a stable configuration is reached. 
When the memory budget for $Q$ is given, we propose using
a removal policy of those elements that are less frequently solutions of eq. (\ref{eq:fullmatch}). 
This resembles somehow curriculum learning~\cite{curriculumlearning}, where examples are presented to learning systems following an increasing degree of complexity. 
Interestingly, the proposed blurring scheme is also related to the process which takes place in infants
during the development of their visual skills~\cite{Darwin-1877,DOBSON-VR-1978,Turkewitz-1982}.
In a sense, the  agent gets rid of the information overload and  operates with the amount of information that it can handle at the current stage of development. Interestingly, this seems to be rooted in information-based principles more than in biology.
%

\marginpar{{\em  Deep nets and \\ developmental stages}} 
At each layer of the deep net, the feature learning is based on the same principles and algorithms. 
However, we use developmental stages based on learning layers separately, so as upper
layers activate the learning process only when the features of the lower layers have already been
learned. The pixel-based features that are developed in the whole net are used for the construction of higher-level
representations that are involved in the prediction of symbolic functions.

\section{Learning symbolic constraints}
\label{sec:symbolic}
%
In order to build high-level symbolic functions, we first aggregate homogenous regions (superpixels) of $\CalV_t$, as described in Section \ref{sec:architecture}. This reduces the computational burden of pixel-based processing, but it requires to move from the pixel-based descriptors $p_c^\ell(x,\CalI_t)$ to region-based descriptors $s_{z_t}$, where $z_t$ is the index of a region of $\CalV_t$.

\marginpar{{\em High-level \\ representations}} 
In detail, the aggregation procedure generates $R$ regions\footnote{In the following description we do not explicit the dependence of the region variables on time $t$ to keep the notation simple.} (superpixels) $r_z$, $z=1,\ldots,R$, where each $r_z$ collects the coordinates of the pixels belonging to the $z$-th region.
While the bare average of $p_c^\ell$ over all the $x \in r_z$ could be directly exploited to build $s_z$, we aggregate $p_c^\ell$ by means of a co-occurrence criterion. This allows us to consider spatial relationships among the pixel features that would be otherwise lost in the averaging process.
First, we determine the winning feature in $p_c^\ell$. Then, we count the occurrences of pairs of winning features in the neighborhood of $x$, for all pixels in the region, building a histogram that is normalized by the total number of counts. 
The normalization yields a representation that is invariant w.r.t. scale changes (number of pixels) of the region $r_z$.
Then, we repeat the process for all the categories and layers, stacking the resulting histograms to generate the region descriptor $s_z$. We also add the color histogram over $r_z$, considering $4$ equispaced bins for each channel of the considered (RGB or Lab) color space.
Finally, we normalize $s_z$ to sum to one, giving the same weight to the feature-based portion of $s_z$ and to the color-based one (more generally, the weight of the two portions could be tuned by a customizable parameter).
The length of $s_z$ is $0.5 \sum_{c,\ell} d_c^\ell (d_c^\ell+1) + 4^{3}$.

Region descriptors and their relationships are stored as vertices (also referred to as ``nodes'') and edges of the \textit{Developmental Object Graph} (DOG).
Nodes are the entities on which a tag-prediction is performed, whereas edges are used to generate coherence constraints, as detailed in the following. Similarly to the case of the set $Q$ in Section \ref{sec:features}, the set $V$ of the nodes in the DOG is an $\epsilon$-net in which the minimum distance between pairs of nodes is $\tau$, a user-defined tolerance.
Each region descriptor $s_z$, is either mapped to its nearest neighbor in $V$ (if the distance from it is $\leq \tau$), or it is added to $V$ (if the distance is greater than $ \tau$). In the former case, we say that ``$s_z$ hits node $v_j$'', and $s_z$ inherit the tag-predictions of $v_j$, that can be easily cached (Section \ref{sec:lfc}). As for the nearest neighbor computations concerning receptive inputs, also in this case the search procedure can be efficiently performed by partitioning the search space, and by tolerating sub-optimal mappings. A pre-defined time budget is defined, crucial for real-time systems, and we return the best response within such time constraint.
The $\chi^2$ distance is exploited, since it is well suited for comparing histograms.

\marginpar{{\em Node \\ construction \\ by motion \\ estimation }} 
As for receptive inputs, we can also use motion coherence to strongly reduce the number of full searches required to map the region descriptors to the nodes of $V$.
We partition the image into $\kappa \times \kappa$ rectangular portions of the same size, and we associate each region to the portion containing its barycenter. Given the region-descriptor-to-node mappings computed in the frame $\CalV_{t-1}$, we can search for valid hits at time $t$ by comparing the current descriptors with the nodes associated to the regions of the nearby image portions in the previous frame.

\marginpar{{\em Spatial and \\ motion-based \\ edges }} 
DOG edges are of two different types, \textit{spatial} and \textit{motion-based}, and their weights are indicated with $w^{s}_{ij}$ and $w^{m}_{ij}$ and stored into the (symmetric) adjacency matrices $W^{s}$ and $W^{m}$, respectively.
Spatial connections are built by assuming that close descriptors represent similar visual patterns. 
Only those nodes that are closer than a predefined factor $\gamma_{s} > \tau$ are connected, leading to a sparse set of edges.
The edge weights are computed by the $\chi^2$ Gaussian kernel, as $w^{s}_{ij} = \exp\left(-\frac{\chi^2(v_i,v_j)}{2\sigma_{\tau}^2}\right)$.

Nodes that represent regions with similar appearance may not be actually spatially close due to slight variations in lighting conditions, occlusions, or due to the suboptimal solutions of the $\xi$ matching process (Section \ref{sec:features}). The motion between frames $\CalV_{t-1}$ and $\CalV_{t}$ can be used to overcome this issue, and, for this reason, we introduce links between nodes that are estimated to be the source and the destination of a motion flow.
The weights are initialized as $w^{m}_{ij}=0$ at $t=0$, for each pair $(i,j)$, and then they are estimated by a two-step process. First the likelihood $P_t(v_a, v_b)$, that two DOG nodes $v_a, v_b \in V$ are related in two consecutive frames $\CalV_{t}$ and $\CalV_{t-1}$ due to the estimated motion, is computed. Then the weight $w^m_{a,b}$ of the edge between the two corresponding DOG nodes  is updated. $P_t(v_a, v_b)$ is computed by considering the motion vectors that connect each pixel in $\CalV_{t}$ to another pixel of $\CalV_{t-1}$ (Section \ref{sec:features}). For each pair of connected pixels, one belonging to region $r_{z_t} \subset \CalV_{t}$ and the other to $r_{h_{t-1}} \subset \CalV_{t-1}$, we consider the DOG nodes $v_a$ and $v_b$ to which $r_{z_t}$ and $r_{h_{t-1}}$ are respectively associated.
The observed event gives an evidence of the link between $v_a$ and $v_b$, and, hence, 
the frequency count for $P_t(v_a, v_b)$ is increased by a vote, scaled by $|r_{z_t}|$ to avoid penalizing smaller regions. Moreover, in the computation we consider only votes involving regions of comparable size, i.e. $\frac{\left|r_{z_t}\right|}{\left|r_{h_{t-1}}\right|}\in \left[0.8, 1.25\right]$, to reduce the effects due to significant changes in the detection of regions in two consecutive frames. Finally, since a DOG node $v$ corresponds to all the region descriptors that hit it,  the total votes accumulated for the edge between $v_a$ and $v_b$ are also scaled by the number of distinct regions of $\CalV_{t}$ that contributed to the votes. Similarly to the spatial case, a sparse connectivity is favored by pruning the estimates below a given threshold $\gamma_{m}$, in order to avoid adding weak connections due to noisy motion predictions. 

The edge weights are computed by averaging in time the computed  $P_t(v_a, v_b)$, as $w^{m}_{a,b} = \frac{1}{t+1} \sum_{u=0}^{t} P_u(v_a, v_b)$. This step can be done with an incremental update that does not require to store the likelihood estimates for all the time steps.




The agent interacts with the external environment, gathering different kinds of knowledge over the data in $V$, represented under the unifying notion of \textit{constraint} \cite{learningfromconstraints} (Section \ref{sec:lfc}). 
The most prominent example of knowledge comes from the interaction with human users (Section \ref{sec:human}) who provide custom class-supervisions (with values in $\{-1,+1\}$ for negative and positive supervision, respectively).
For each DOG node, the agent will be able to predict tag-scores for those classes for which it received at least one (positive) supervision.

At a given time $t$, let us suppose that the agent received supervisions for a set of $\omega$ classes.
We indicate with $f_k$ the function that models the predictor of the $k$-th class, and, again, we follow the framework of Section \ref{sec:lfc}.
We select the $\chi^2$ Gaussian kernel and we also assume that the agent is biased towards negative predictions, i.e. we added a fixed bias term in eq. (\ref{eq:solution}) equal to $-1$, allowing it to learn from positive examples only. 

The constraints in $\CalC$ are of two types: supervision constraints $\mu_{\CalS}^{(1)}$, and coherence constraints $\mu_{\CalM}^{(2)}$.
\marginpar{{\em Supervision \\ constraints}} 
The former enforce the fulfillment of labels $y_{i,k} \in \{-1,+1\}$ on some DOG nodes $v_i \in V$ and for some functions $f_k$. For each $f_k$, the supervised nodes are collected into the set $\CalS_k = \{(v_i,y_{i,k}),\ i = 1,\ldots,l_k \}$, and 
\begin{equation}
\mu_{\CalS}^{(1)}(f) =  \sum_{k=1}^{\omega} \sum_{(v_i,y_{i,k}) \in \CalS_k} \hskip -4mm  \beta_{ik}  \max (0, 1-y_{i,k} f_k(v_i))^2 \ .
\label{eq:sup}
\end{equation}
The scalar $\beta_{ik} > 0$ is the \textit{belief} \cite{learningfromconstraints} of each point-wise constraint. When a new constraint is fed by the user, its belief is set to a fixed initial value. Then, $\beta_{ik}$ is increased if the user provides the same constraint multiple times or decreased in case of mismatching supervisions, keeping $\sum_i \beta_{ik} = 1$. This allows the agent to better focus on those supervisions that have been frequently provided, and to give less weight to noisy and incoherent labels.
Weak supervisions (Section \ref{sec:architecture}) on the tag $k$ are converted into constraints as in eq. (\ref{eq:sup}) by determining if there exists a node associated to the current frame for which the $k$-th tag-score is above a predefined threshold.

\marginpar{{\em Spatial \\ and motion \\ coherence \\ constraints}} 
The coherence constraints enforce a smooth decision over connected vertices of the DOG,
\begin{eqnarray}
\mu_{\CalM}^{(2)}(f) = \sum_{k=1}^{\omega} \sum_{i=1}^{|V|}\sum_{j=i+1}^{|V|} w_{ij} (f_k(v_i) - f_k(v_j))^2 \ ,
\end{eqnarray}
leading to an instance of the classical manifold regularization \cite{melacci2011laplacian}.
In this case, the \textit{belief} of each point-wise constraint is $w_{ij}$, that is given by a linear combination of the aforementioned edge weights $w^{s}_{ij}$ and $w^{m}_{ij}$,
\begin{equation}
w_{ij} = \lambda_{\CalM} \left(\alpha_{\CalM} \cdot  w^{s}_{ij} + (1-\alpha_{\CalM}) \cdot  w^{m}_{ij} \right) \ .
\end{equation}
Here $\lambda_{\CalM} > 0$ defines the global weight of the coherence constraints while  $\alpha_{\CalM} \in [0,1]$ can be used to tune the strength of the spatial-based connections w.r.t. the motion-based ones.

Following Section \ref{sec:lfc}, the solution of the problem is given by eq. (\ref{eq:solution}) and it is a kernel expansion $f_k (v) = \sum_{i=1}^{|V|} o_{ik} k(v_i, v)$, where $v_{i}$ is the real valued descriptor associated to the corresponding DOG node in $V$. 
The set of DOG nodes $V$ progressively grows as the video stream is processed, up to a predefined maximum size (due to the selected memory budget). 
When the set $V$ reaches the maximum allowed size, the adopted removal policy selects those nodes that have not been recently hit by any descriptors, with a small number of hits, and that are not involved by any supervision constraint.




\section{Experiments}
\label{sec:exp}
In this section, we present experimental results to evaluate several aspects of the DVA architecture, from feature extraction up to the symbolic level of semantic labeling. Experiments were carried out on a variety of different videos ranging from artificial worlds and cartoons to real-world scenes, to show the flexibility of learning in  unrestricted visual environments. The website of the project (\url{http://dva.diism.unisi.it}) hosts supplementary material with video sequences illustrating several case studies. The DVA software package can also be downloaded and installed under different versions of Linux and Mac OS X. 

\subsection{Feature extraction}
\label{sec:expfeatures}
\begin{figure}
\centering
\includegraphics[width=0.7\textwidth]{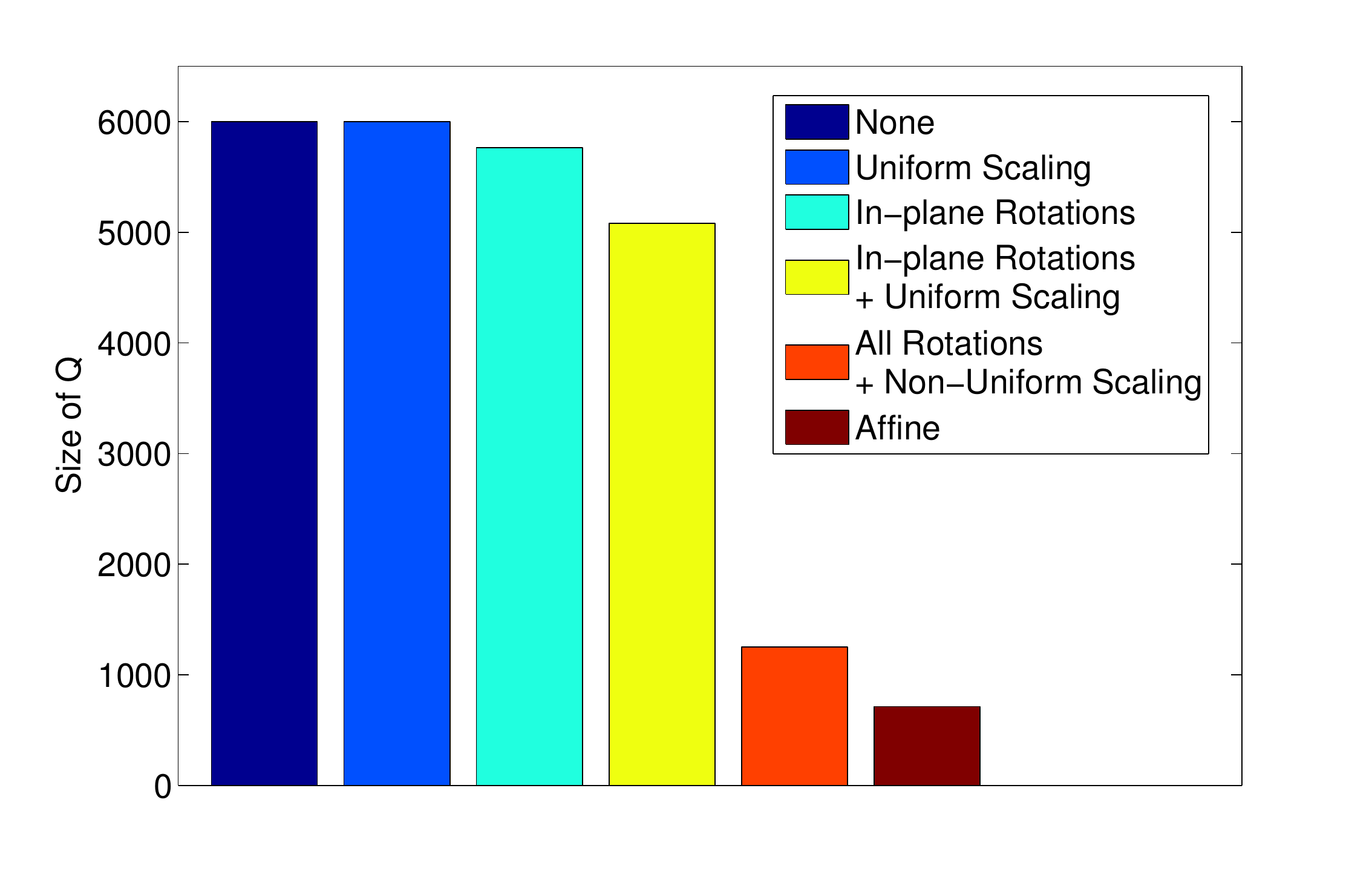} 
\vskip -3mm
\caption{The size of $Q$ (Section \ref{sec:features}) when different invariances to geometric transformations are activated. In these experiments, the memory budget was set to 6,000 data points.}
\label{fig:Q_size}
\end{figure}

\begin{figure*}
\centering
{\footnotesize \textsc{Frame} \hskip 2.6cm \textsc{Layer 0} \hskip 2.5cm \textsc{Layer 1} \hskip 2.3cm \textsc{Layer 2} }\\
\vskip -3mm $\ $ \\
\includegraphics[width=0.24\textwidth]{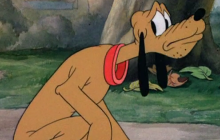} 
\includegraphics[width=0.24\textwidth]{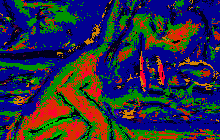} 
\includegraphics[width=0.24\textwidth]{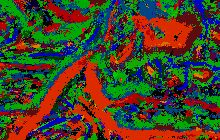} 
\includegraphics[width=0.24\textwidth]{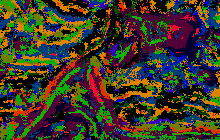} \\
\vskip -3mm $\ $ \\
\includegraphics[width=0.24\textwidth]{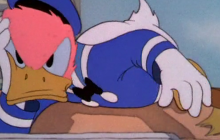} 
\includegraphics[width=0.24\textwidth]{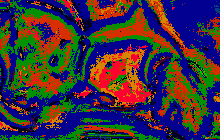} 
\includegraphics[width=0.24\textwidth]{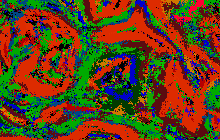} 
\includegraphics[width=0.24\textwidth]{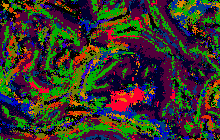} \\
\vskip -3mm $\ $ \\
\includegraphics[width=0.24\textwidth]{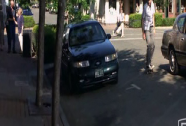} 
\includegraphics[width=0.24\textwidth]{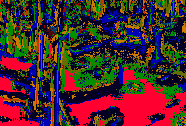} 
\includegraphics[width=0.24\textwidth]{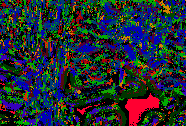} 
\includegraphics[width=0.24\textwidth]{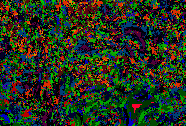} \\
\vskip -3mm $\ $ \\
\includegraphics[width=0.24\textwidth]{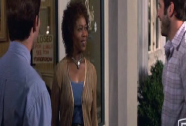} 
\includegraphics[width=0.24\textwidth]{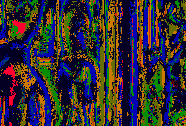} 
\includegraphics[width=0.24\textwidth]{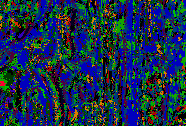} 
\includegraphics[width=0.24\textwidth]{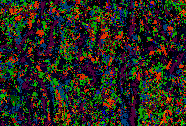} \\
\caption{The feature maps of a 3 layered DVA, processing a cartoon clip (Donald Duck, \textcopyright\ The Walt Disney Company) and a sequence from the Hollywood Dataset HOHA 2 \cite{hoha2}. Each pixel is depicted with the color that corresponds to the winning feature (best viewed in color).}
\label{fig:feature_maps}
\end{figure*}

We evaluated the impact of the invariances in building the  set $Q$ from which the low-level features 
are learned. A shallow DVA (1 layer) was run on three unrelated real-world video sequences from the Hollywood Dataset HOHA 2 \cite{hoha2}, so as to explore the growth of  $Q$. 
Videos were rescaled to $320 \times 240$, and converted to grayscale.
We selected an architecture with $\CalN = 5\times 5$ receptive fields, $\epsilon = 0.7$, and we repeated the experiment by activating invariances to different classes of geometric transformations (with $|\Phi_1|=16$, $|\Phi_2|=3$, $|\Phi_3|\leq6$, $|\Sigma|=3$, see Section \ref{sec:features}).
Figure \ref{fig:Q_size} highlights the crucial impact of using invariances for reducing $|Q|$. We set a memory budget that allowed DVA to store up to 6,000 $\xi$'s into $Q$. When full affine invariance is activated, there is a significant reduction of $|Q|$, thus simplifying the feature learning procedure. When considering the case with no-invariances, we reached the budget-limit earlier that in the case of scale-invariance-only.
A deeper DVA (3 layers) processed the same sequences in order to learn $d^{\ell}=20$ features per layer, $\ell=1,2,3$ (one-category, $C^{\ell}=1$). The same architecture was also used in processing a cartoon clip with different resolution.
Figure \ref{fig:feature_maps} shows the feature maps on four sample frames. Each pixel is depicted with the color that corresponds to the winning feature, i.e. the color of $x$ is indexed by $\arg\max_j \{p_{j}^{\ell}(x,\cdot),\ j=1,\ldots,d^{\ell}\}$. While features of the lowest layer easily follow the details of the input, higher layers develop functions that capture more abstract visual patterns.
From the third row of Figure \ref{fig:feature_maps}, we can see that bright red pixels basically indicate the feature associated with constant receptive inputs. Moving toward the higher layers,  such a feature becomes less evident, since the hierarchical application of the receptive fields virtually captures larger portions of the input frame, thus reducing the probability of constant patterns. 
From  the last two rows of Figure \ref{fig:feature_maps}, 
the orange feature seems to capture edge-like patterns,  independently of their orientation, 
thanks  to the invariance property of the DVA features. For instance, we can appreciate that such feature is high both along vertical 
and horizontal edges.
Notice that feature orientation, scale, and  other transformation-related properties are defined by the heuristic searching procedure of Section \ref{sec:features}. Hence, for each pixel,
these transformations can also be recovered.


\subsection{The role of motion}
\label{sec:expmotion}
Motion plays a crucial role at several levels of the DVA architecture. In Section \ref{sec:features} we have
seen that handling invariances to geometric transformations allows DVA to estimate the motion. 
It turns out that, while imposing motion coherence on low-level features, the velocity of each pixel is
itself properly determined.

\begin{figure}
\includegraphics[trim=12 12 12 12,clip,width=0.33\textwidth]{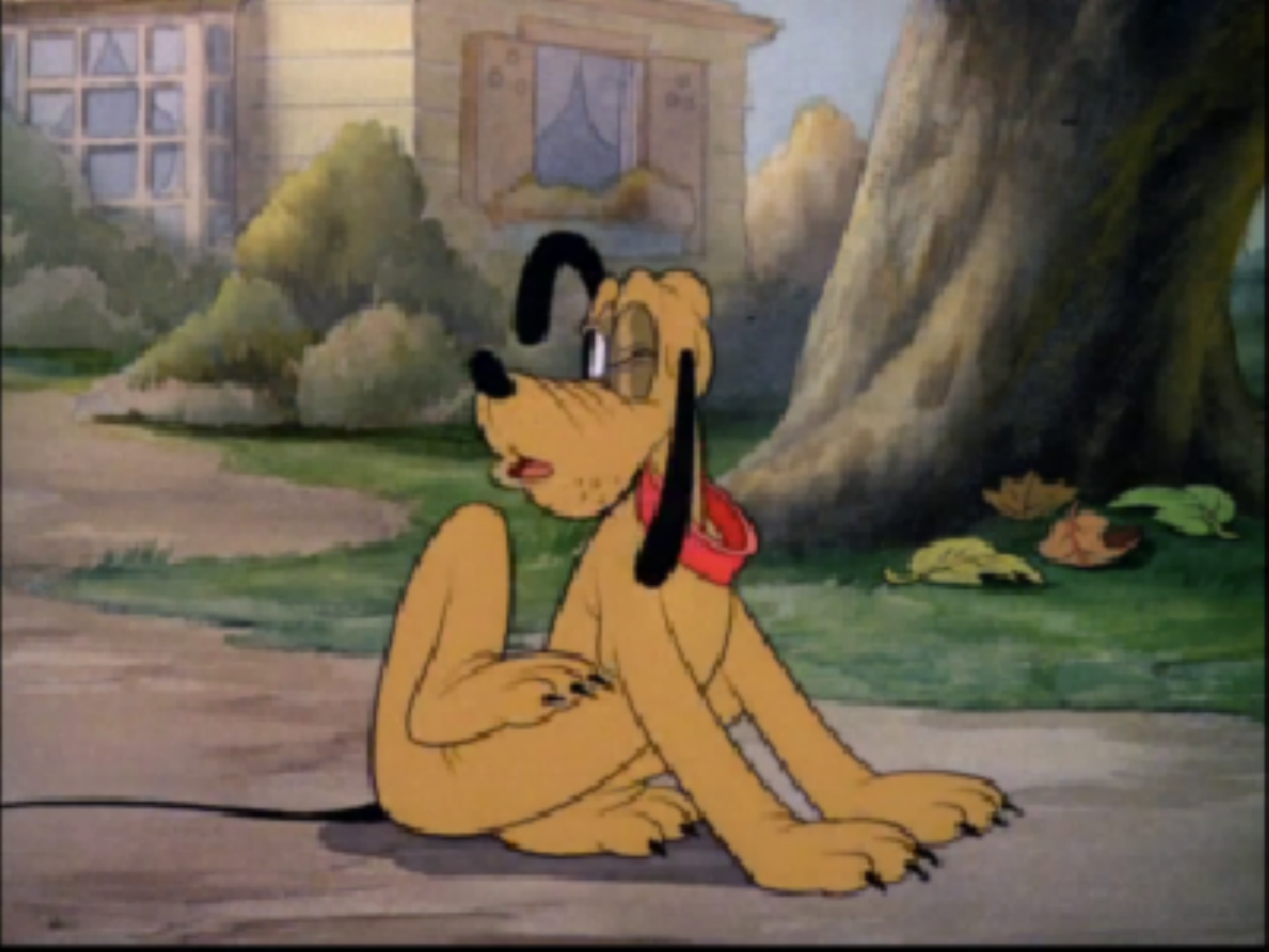}
\includegraphics[trim=12 12 12 12,clip,width=0.33\textwidth]{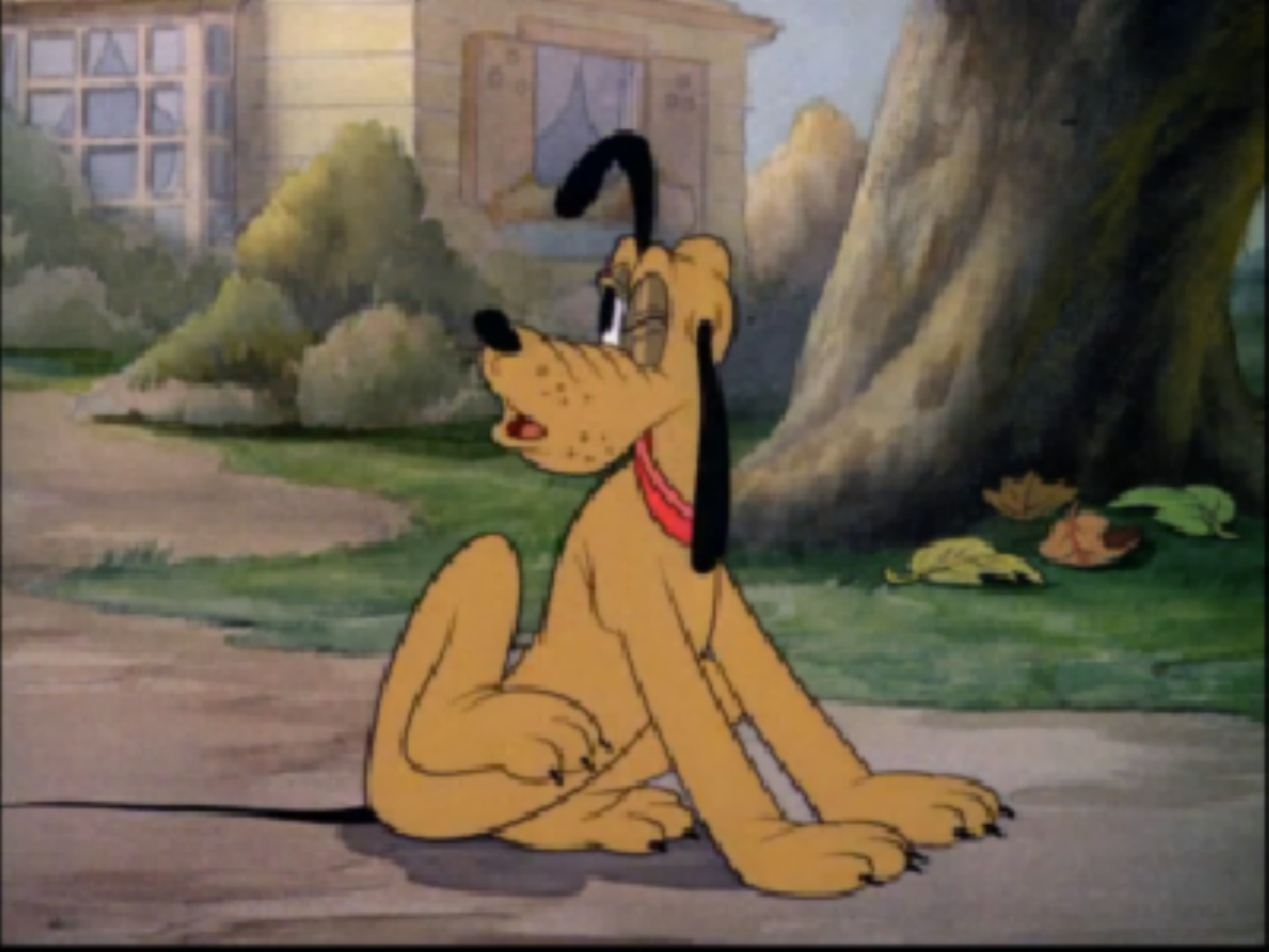}
\includegraphics[width=0.33\textwidth]{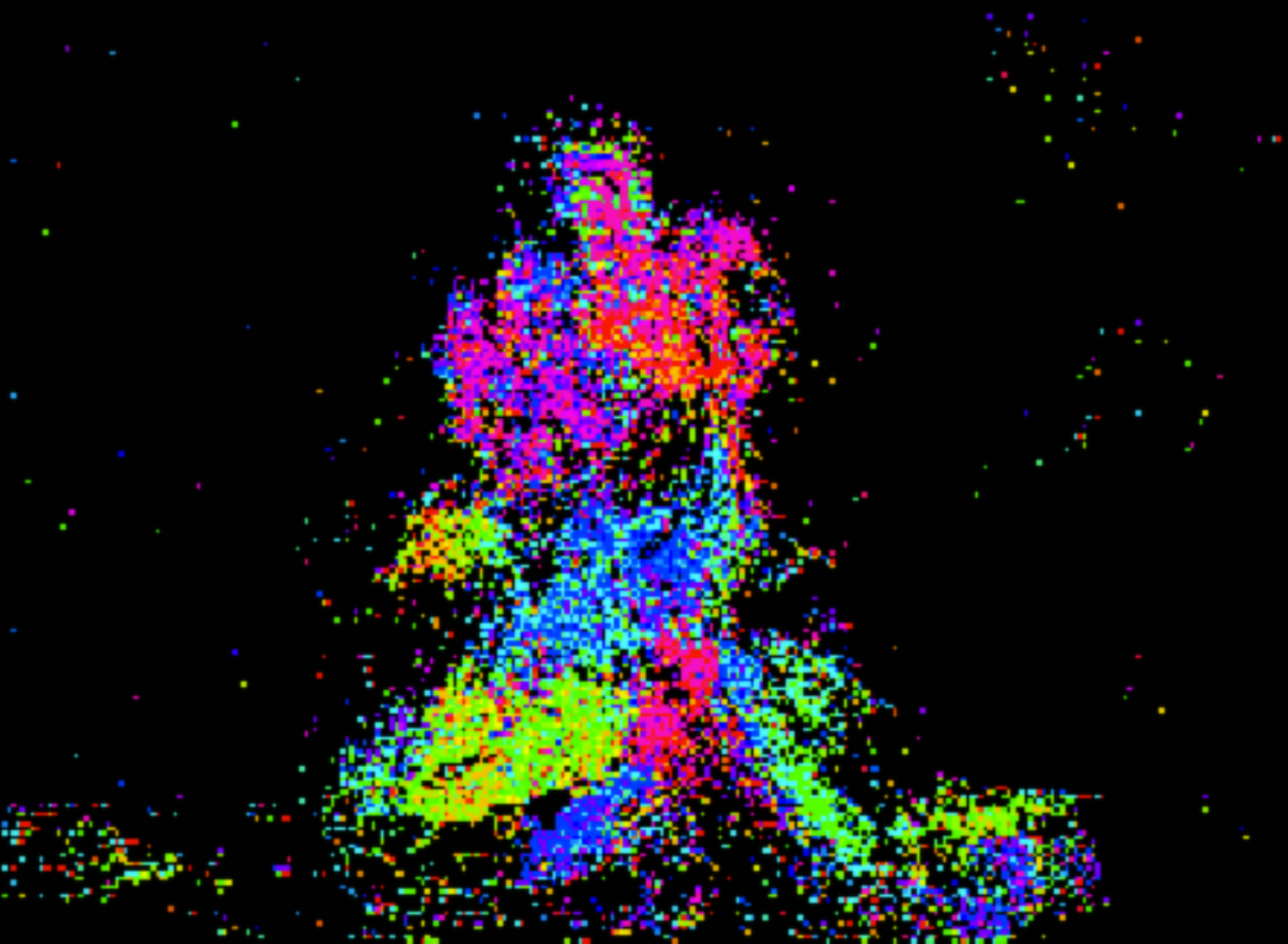}\\
\includegraphics[trim=18 18 18 18,clip,width=0.33\textwidth]{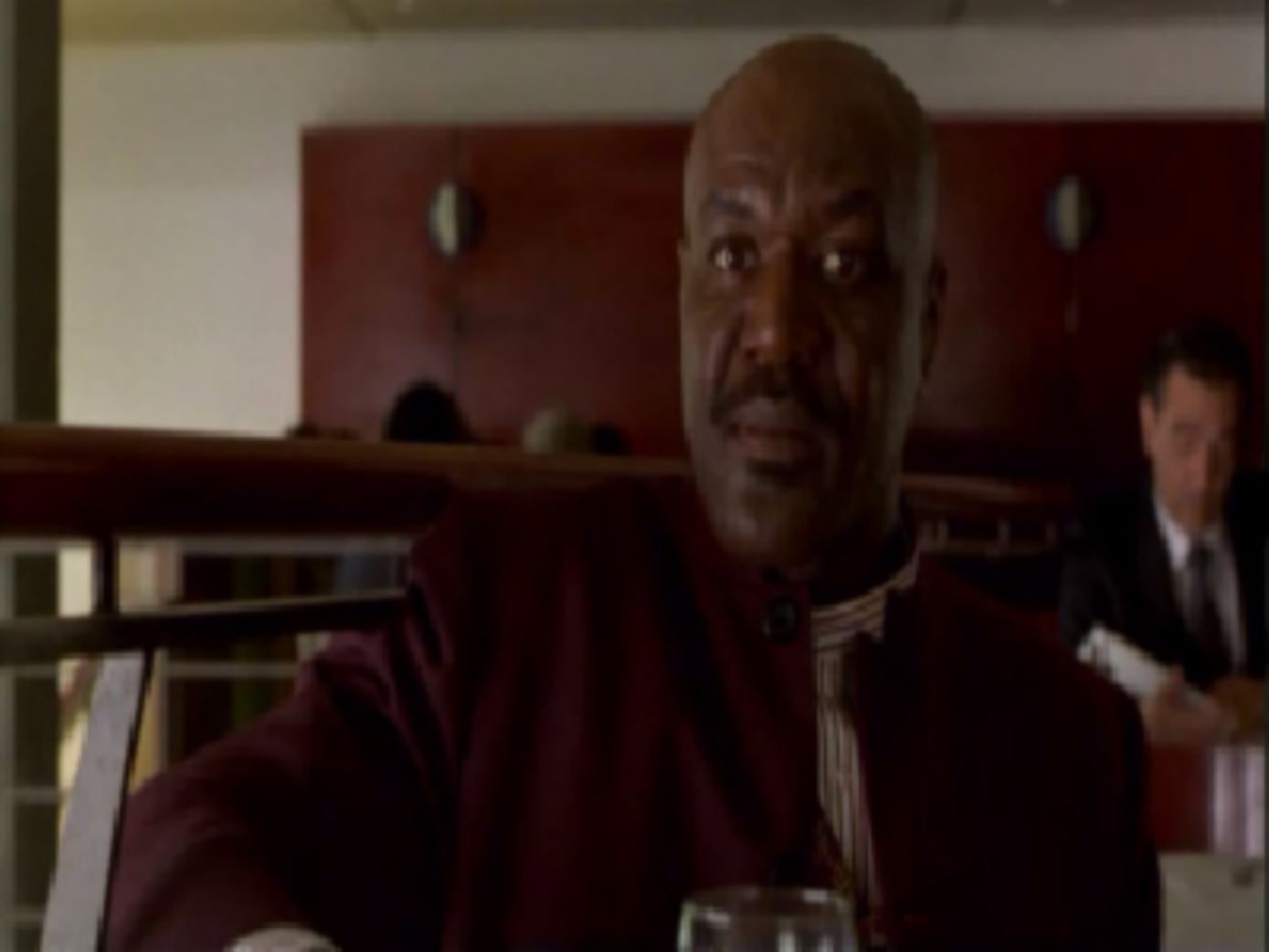}
\includegraphics[trim=18 18 18 18,clip,width=0.33\textwidth]{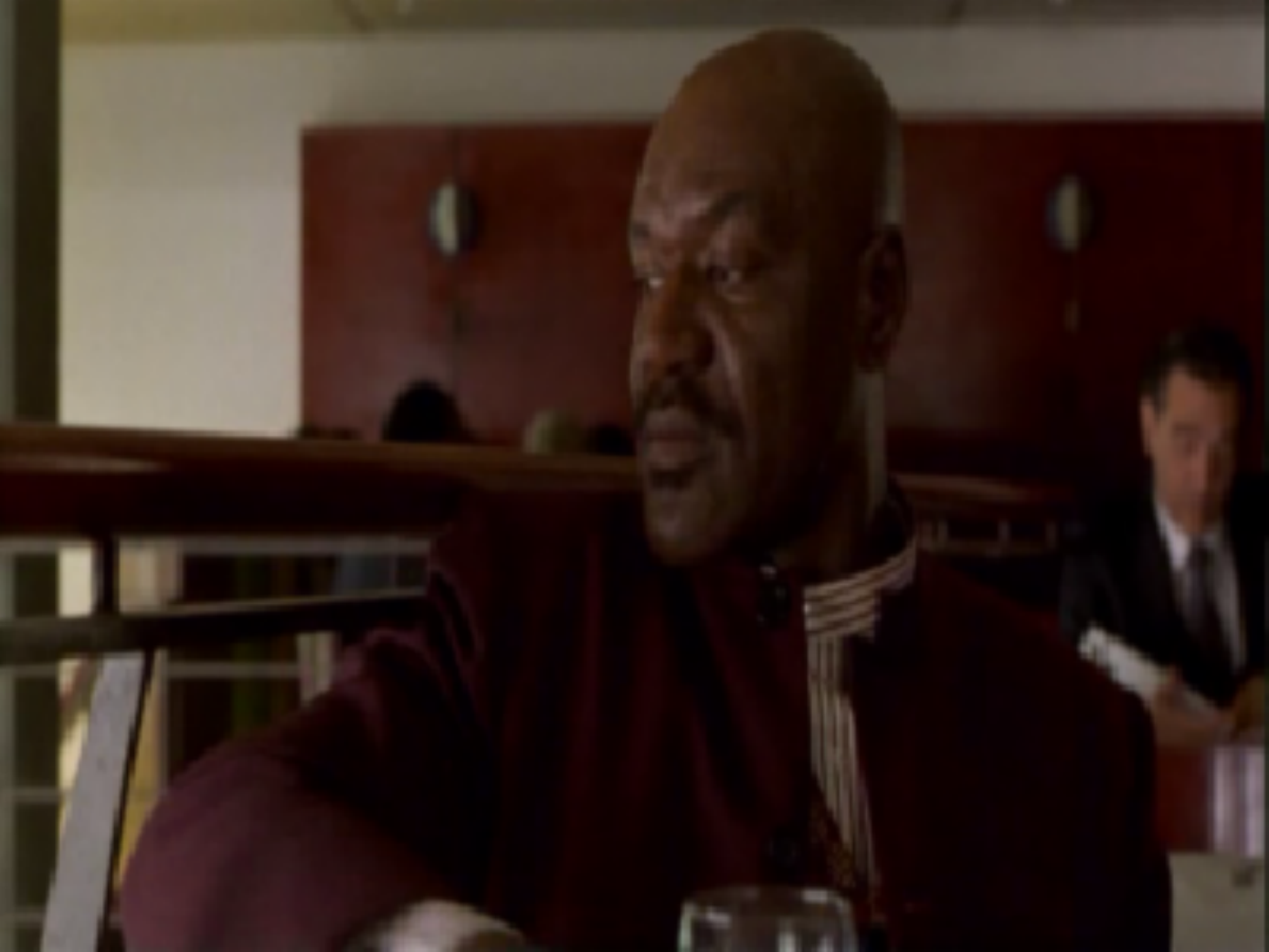}
\includegraphics[width=0.33\textwidth]{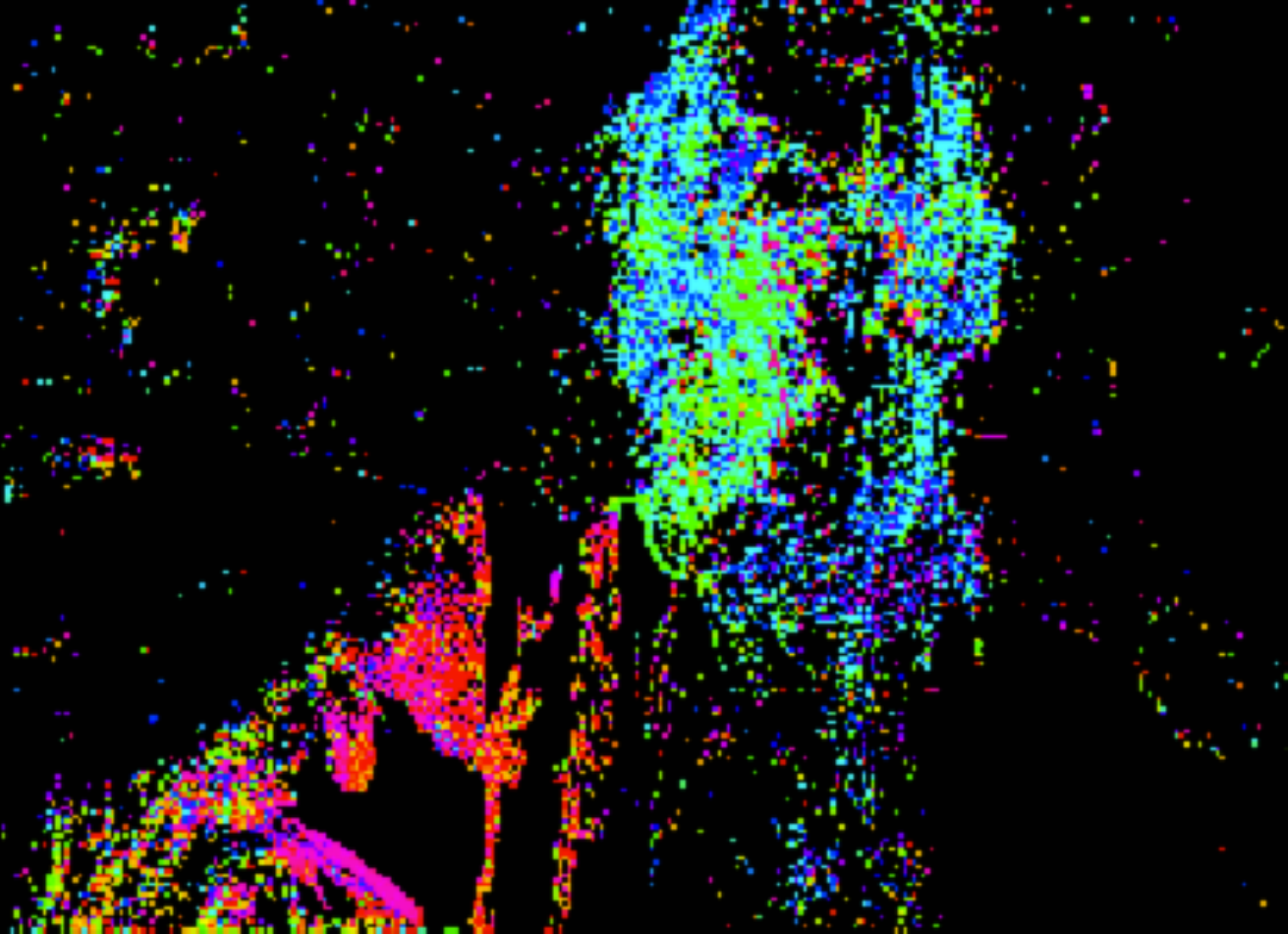}
\caption{Two examples of motion estimation. Two subsequent frames are presented (column 1 and 2) 
along with a colormap of motion estimation per pixel (column 3), where colors indicate 
the angle of the estimated velocity vector, according to the hue colorwheel (where red equal to $0$ degrees). Top row: The dog is lowering its leg (green/yellow) while moving its body up (cyan/blue) and its face and ear towards up-right (violet/red) (Donald Duck, \textcopyright\ The Walt Disney Company). Bottom row: the actor is turning his head towards the left edge of the frame (cyan) while moving his right arm towards his chest (red).\label{fig:opticalflow}}
\end{figure}

An example of motion estimation is given in Figure~\ref{fig:opticalflow}, where in the third column each pixel is colored with a different hue according to the angle associated to its velocity vector. In this context, the use of motion coherence in feature extraction is crucial also to speed up computation to solve eq. (\ref{eq:fullmatch}). We used again three random clips from the HOHA 2 dataset, and measured the average computational time required by a 1-layer DVA to process one frame at $320 \times 240$ resolution. The impact of motion is dramatic, as  the required time dropped from 5.2 seconds per frame to 0.3 seconds per frame on an Intel Core-i7 laptop. It is worth 
mentioning that time budget requirements can also be imposed in order to further speed up the computation and perform real-time frame processing.

\begin{figure}
\begin{center}
\includegraphics[width=0.48\textwidth]{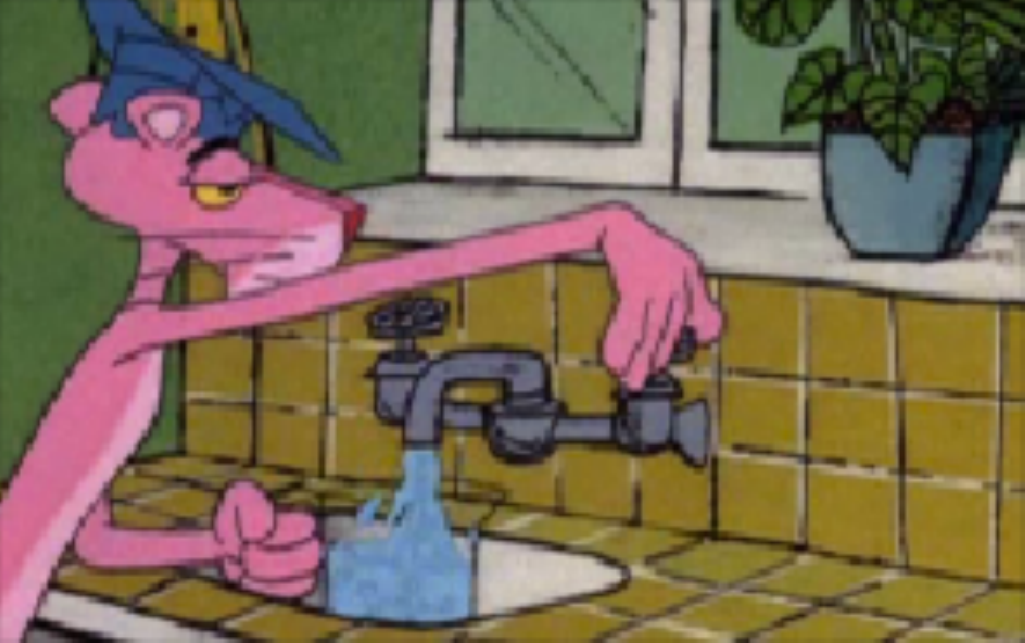}
\includegraphics[width=0.48\textwidth]{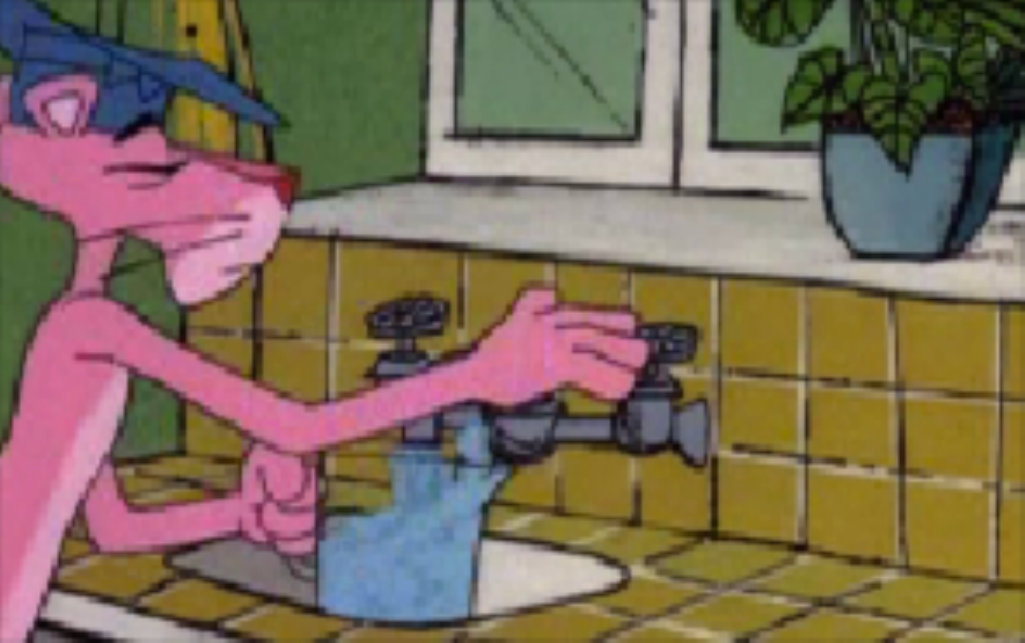}
\\\vspace{0.1cm}
\includegraphics[width=0.48\textwidth]{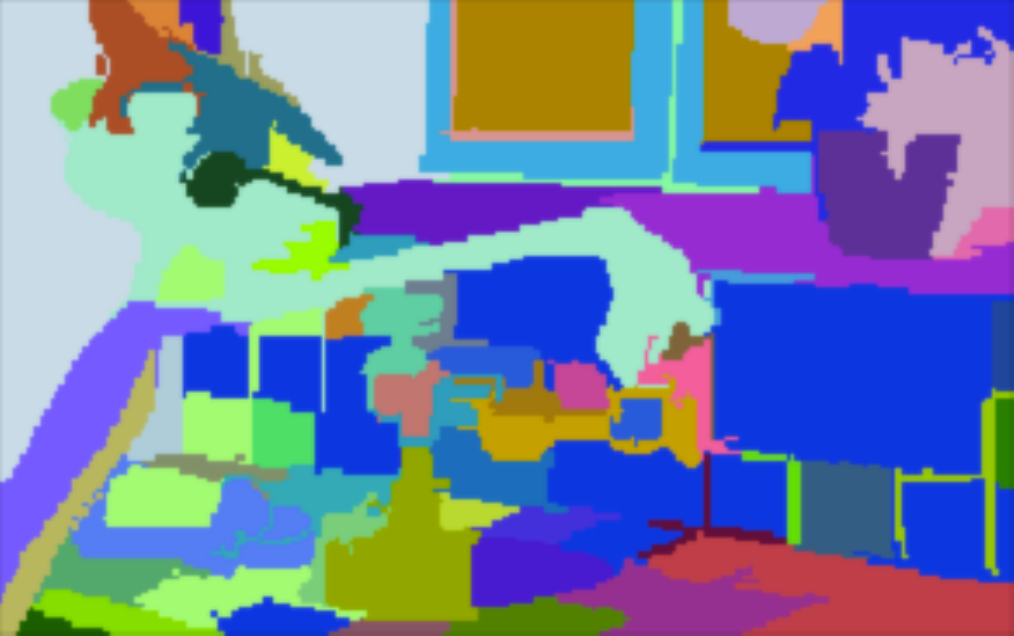}
\includegraphics[width=0.48\textwidth]{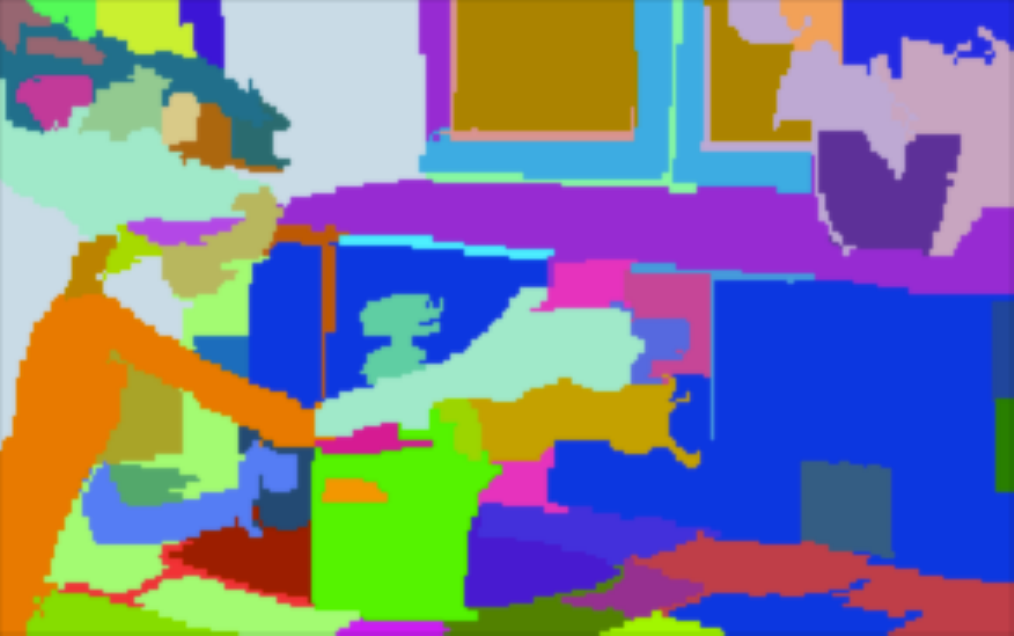}
\\\vspace{0.1cm}
\includegraphics[width=0.48\textwidth]{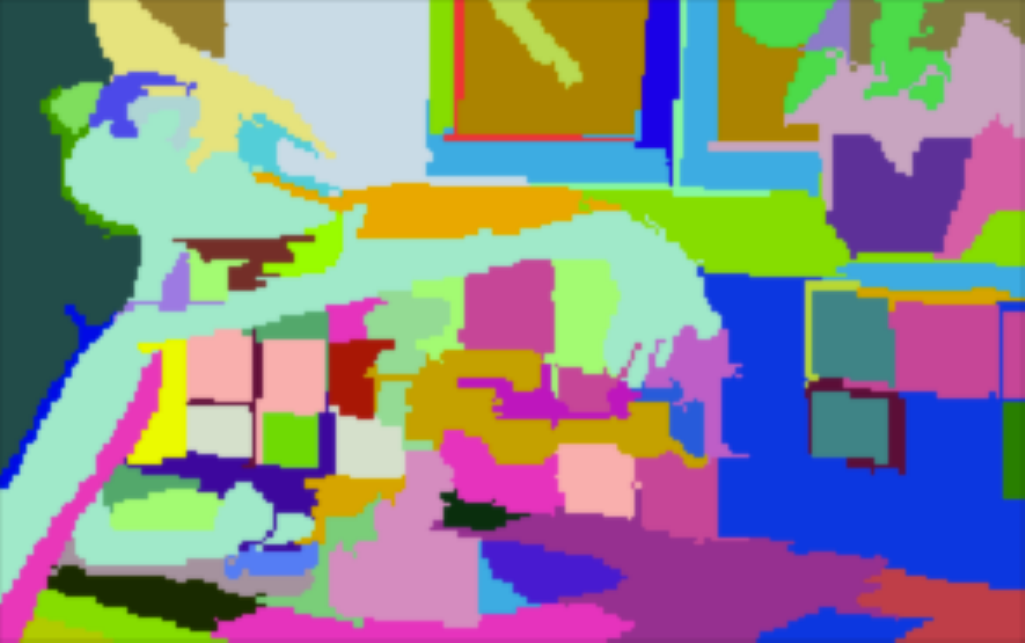}
\includegraphics[width=0.48\textwidth]{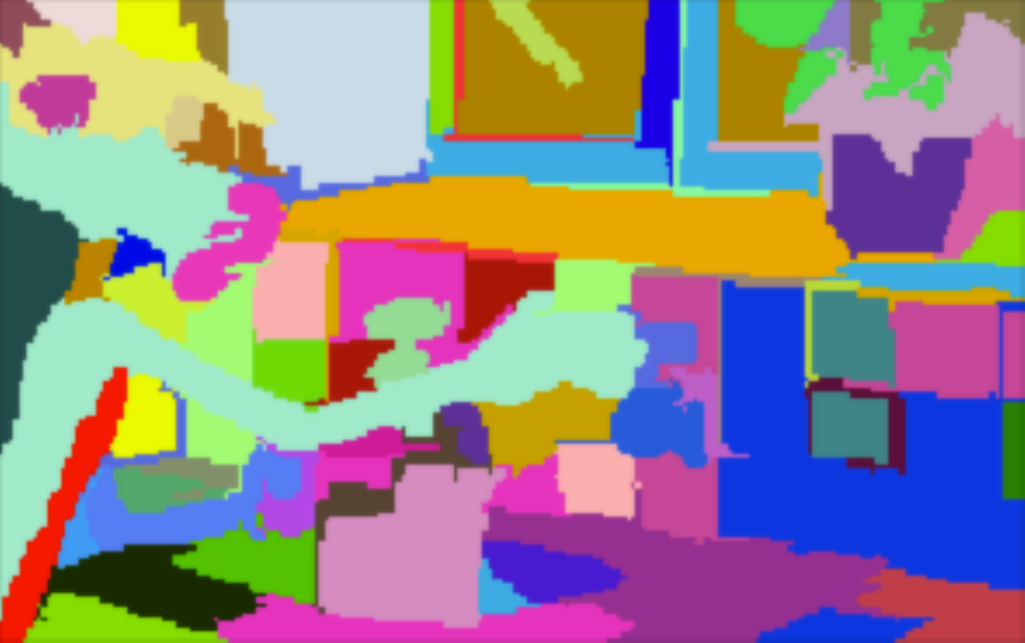}
\caption{The effect of motion coherence on aggregation. Top: two consecutive frames in a Pink Panther cartoon (\textcopyright\ Metro Goldwyn Mayer); middle/bottom: aggregation without/with motion coherence. The same color corresponds to the same node within the DOG, which means identical descriptors, up to tolerance $\tau$. More nodes are kept by exploiting motion coherence (e.g., the body of the Pink Panther, the water carafe, the faucet and tiles).\label{fig:segmentation}}
\end{center}
\end{figure}

Now we show some results illustrating the impact of using motion coherence in the process of region aggregation. Figure~\ref{fig:segmentation} shows a pair of consecutive frames taken from a Pink Panther cartoon (top row), with the results obtained by the region-growing algorithm without (middle row) or with (bottom row) exploiting motion coherence. Regions having the same color are mapped to the same DOG node, therefore sharing very similar descriptors (their distance being $\leq \tau$, see Section~\ref{sec:symbolic}). This example shows that the role of motion is crucial in order to get coherent regions through time
\footnote{Clearly, this simplifies dramatically the subsequent recognition process.}. 
In the middle row we can observe that the body of the Pink Panther changes from blue to orange, while it is always light green when exploiting motion information (bottom row); the water carafe, the faucet, 
and the tiles in the central part of the frame are other examples highlighting this phenomenon.


\begin{figure}
\begin{center}
\setlength{\fboxsep}{1pt}
\setlength{\fboxrule}{2pt}
\fbox{\includegraphics[width=0.16\textwidth]{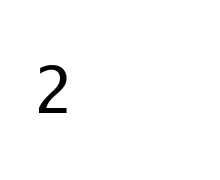}} \hskip 1mm
\fbox{\includegraphics[width=0.16\textwidth]{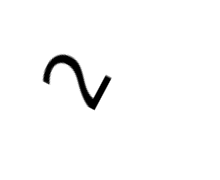}} \hskip 1mm
\fbox{\includegraphics[width=0.16\textwidth]{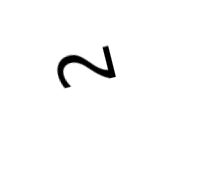}} \hskip 1mm
\fbox{\includegraphics[width=0.16\textwidth]{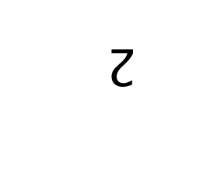}} \hskip 1mm
\fbox{\includegraphics[width=0.16\textwidth]{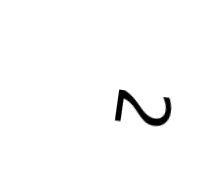}} 
\caption{The {\em digits visual environment}. Each digit moves from left to right by
translations and rotations. While moving, it also scale up and down, as depicted for digit ``2''.
DVAs are required to learn the class for any pixel and any frame.}
\label{fig:digits}
\end{center}
\end{figure}

\marginpar{{\em Digit test: \\ learning to see \\ with one \\ supervision}} 
In order to investigate the effect of motion constraints in the development of the symbolic functions (Section \ref{sec:symbolic}), we created a visual environment of Lucida (antialiased) digits
which move from the left to right by generic roto-translations. While moving, the digits also scale up and down. 
Each digit (from ``0'' to ``9'') follows the same trajectory, as shown 
in Figure \ref{fig:digits} for the case of digit ``2''. 
The visual environment consists of a (loop) collection of 1,378 frames 
with resolution of $220 \times 180$ pixels, 
with a playback speed of $10$ frames per second.
A DVA processed the visual environment while a human supervisor interacted with the agent
by providing only 11 pixel-wise positive supervisions, i.e. 1 per-digit and 1 for the background.
The descriptors of rotated/scaled instances of the same digit turned out to be similar, 
due to the invariance properties of the low-level features.
On the other hand, no all descriptors were mapped to the same DOG node, 
since when imposing memory and time budget, 
the solution of eq. (\ref{eq:fullmatch}) by local coherence (Section \ref{sec:features}) 
might yield suboptimal results. 
Because of the simplicity of this visual environment, 
we selected a shallow DVA architecture, and we kept a real-time processing of the video: $5 \times 5$ receptive fields, with minimum $\sigma$ equal to $5$, and $50$ output features.
We compared three different settings to construct the symbolic functions: the first one is based on supervision constraints only; the second one adds spatial coherence constraints; the last one includes motion coherence constraints too. We also evaluated a baseline linear SVM that processed the whole
frames, rescaled to $44 \times 36$, while each pixel-wise supervision was associated to the corresponding
frame. For this reason, when reporting the results, we excluded the background class that cannot be 
predicted by the baseline SVM, while DVA can easily predict the background just like any other class.
We also generated negative supervisions, that were not used by the DVA, 
to train the SVM in a one-vs-all scheme. Table \ref{tab:digits_results} reports the macro accuracy for the digit-classes (excluding class of digit ``9'', that is not distinguishable from a rotated instance of digit ``6'').
\begin{table}
\centering
\begin{tabular}{l|r}
Model & Accuracy \\
\hline
SVM classifier (baseline) & 7.99\%\\ 
DVA, 10 sup. constraints & 87.47\%\\ 
DVA, 10 sup. + spatial coherence constr. & 92.70\%\\ 
DVA, 10 sup. + spatial/motion coherence constr. & \textbf{99.76}\% 
\end{tabular}
\caption{Macro accuracy on the {\em digit visual environment}. 
Notice that these experiments consider the extreme case in which each class received one supervision only. 
Motion coherence constraints play a fundamental role to disentangle the ambiguities among similar digits.}
\label{tab:digits_results}
\end{table}
Clearly, the SVM classifier,  which uses full-frame supervisions only, does generalize 
in the digit visual environment, whereas the DVA produces very good predictions
even with one supervision per class only. 
Spatial coherence constraints allows the DVA to better generalize the prediction on unlabeled DOG nodes, thus exploiting the underlying manifold of the descriptor space. However, 
it turns out that the classes ``2'' and ``5'' are confused, due to the invariance properties of the low-level features that 
yield spatially similar descriptors for these classes. When introducing motion-based constraints 
the DVA  disentangles these ambiguities, since the motion flow enforces a stronger coherence 
over those subsets of nodes that are related to the same digit. Notice that
the enforcement of motion constraints is not influenced by the direction;
the movement  on different trajectories (e.g., playing the video frames in reverse order) 
generates the same results of Table~\ref{tab:digits_results}.

\begin{figure}
\begin{center}
\includegraphics[width=0.24\textwidth]{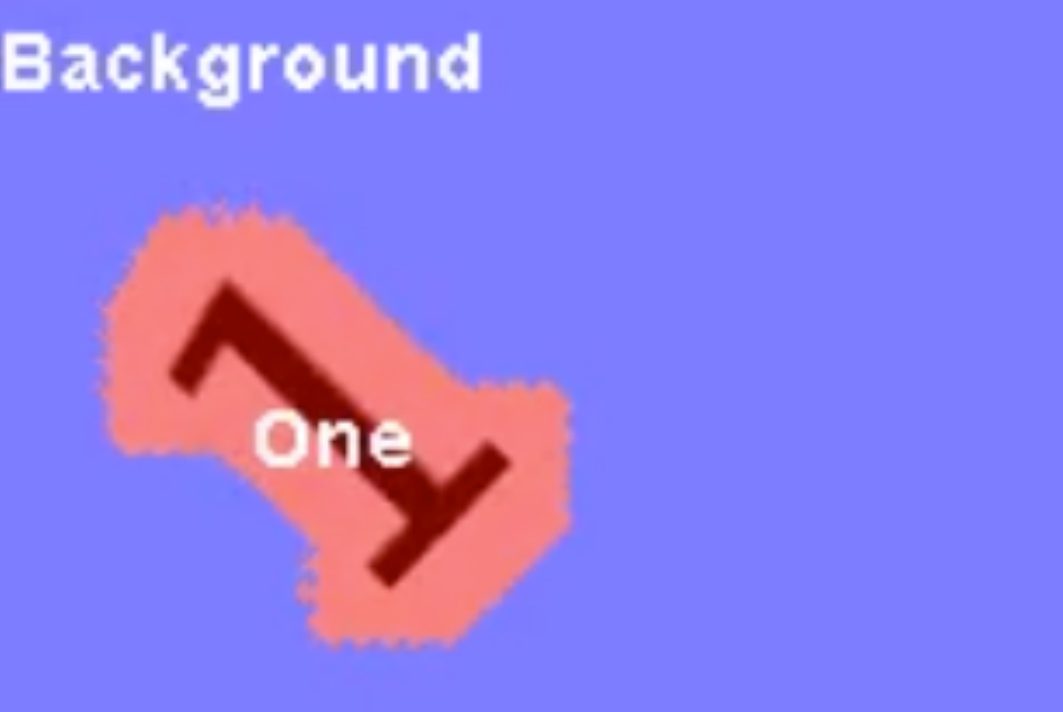} \hskip -0.8mm
\includegraphics[width=0.24\textwidth]{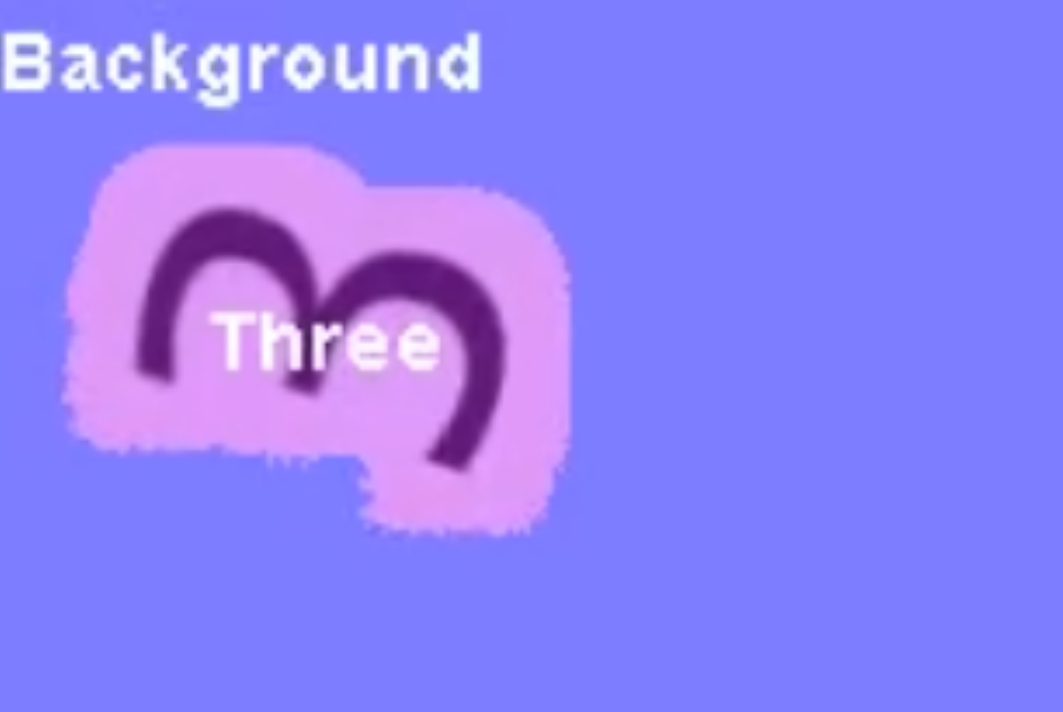} \hskip -0.8mm
\includegraphics[width=0.24\textwidth]{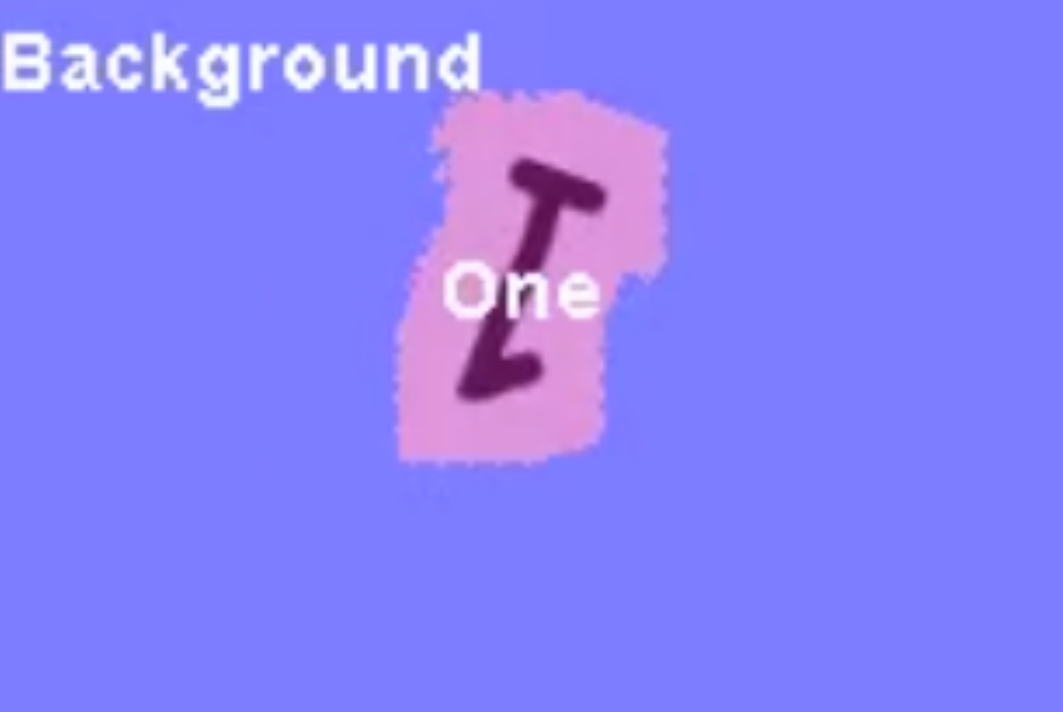} \hskip -0.8mm
\includegraphics[width=0.24\textwidth]{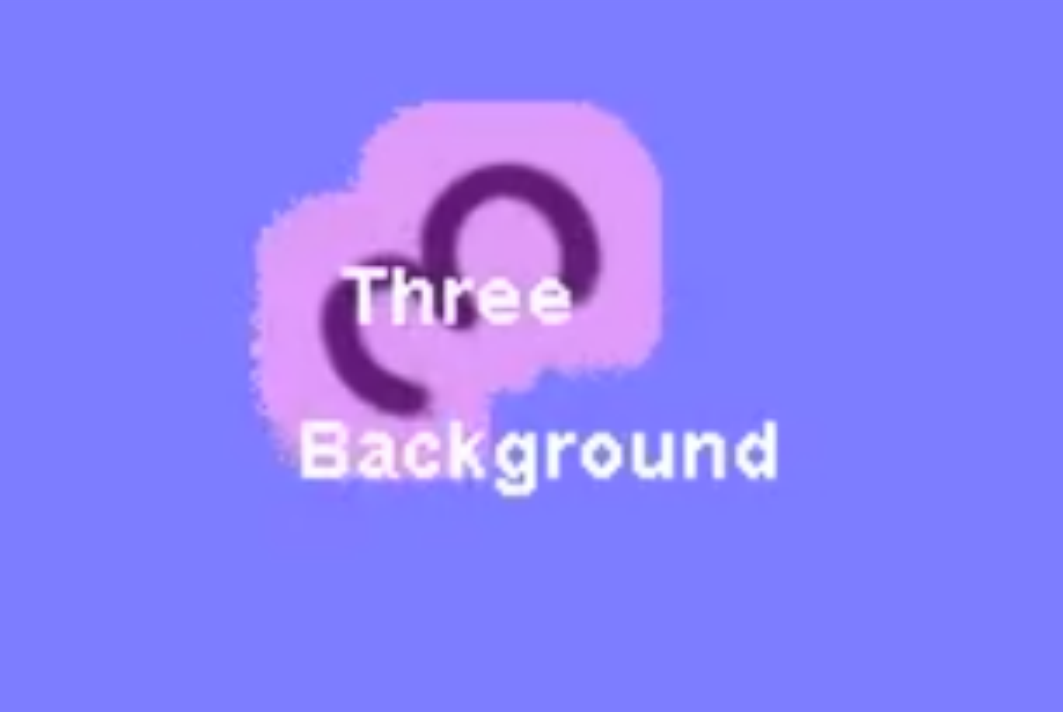}
\caption{Generalization on a different digit visual environment: Lucida (left) vs. Comic Sans MS (right). Only one supervision per digit on the Lucida font was given.\label{fig:digits_font}}
\end{center}
\label{DigitComparison}
\end{figure}

%
%
The experimental results reported for the DVA refer to the case in which it processes 
the given visual environment without making any distinction between learning and test.
The results shown in Table~\ref{tab:digits_results} refer to a configuration in
which the overall structure of the DVA does not change significantly as time goes by. 
We also construct an analogous artificial dataset by using Comic Sans MS font instead of Lucida. 
As shown in Figure~\ref{DigitComparison},
the test of the agent on this new digit visual environment, without supplying any additional supervision, 
yielded very similar results also when playing the video in reverse order.

\subsection{Crowd-sourcing and data-base evaluation}
\label{sec:exppredictions}
DVAs can naturally be evaluated within the L2SLC protocol by crowd-sourcing at
\url{http://dva.diism.unisi.it/rating.html}. In this section we give insights on the
performance of DVAs on some of the many visual environments that we have been
experimenting in our lab. 

\begin{figure*}
\centering
\includegraphics[width=0.24\textwidth]{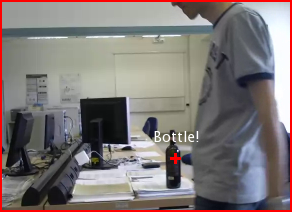}
\includegraphics[width=0.24\textwidth]{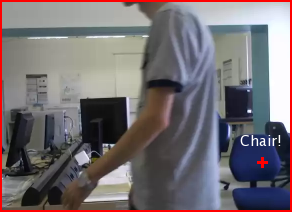}
\includegraphics[width=0.24\textwidth]{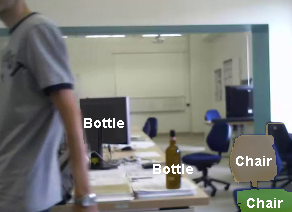}
\includegraphics[width=0.24\textwidth]{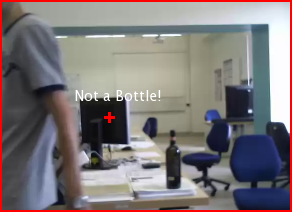}\\
\vskip -3mm $\ $ \\
\includegraphics[width=0.24\textwidth]{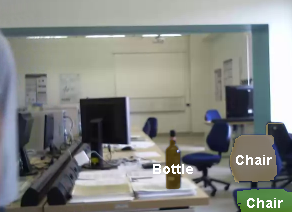}
\includegraphics[width=0.24\textwidth]{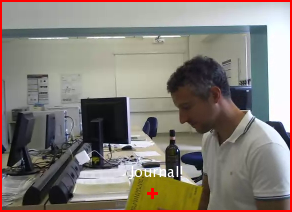}
\includegraphics[width=0.24\textwidth]{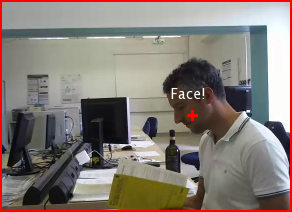}
\includegraphics[width=0.24\textwidth]{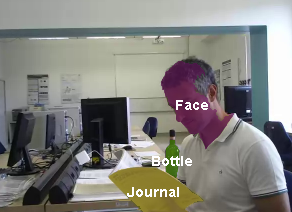}\\
$\ $\\
\vspace{-2mm}$\ $\\
\includegraphics[width=0.24\textwidth]{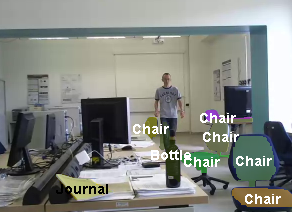}
\includegraphics[width=0.24\textwidth]{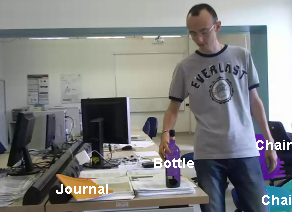}
\includegraphics[width=0.24\textwidth]{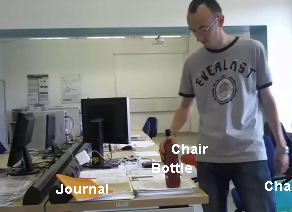}
\includegraphics[width=0.24\textwidth]{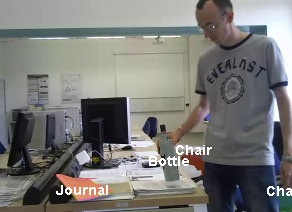}\\
\vskip -3mm $\ $ \\
\includegraphics[width=0.24\textwidth]{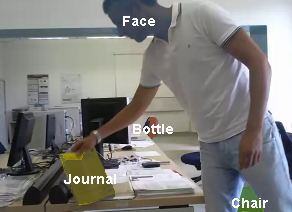}
\includegraphics[width=0.24\textwidth]{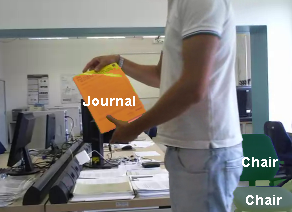}
\includegraphics[width=0.24\textwidth]{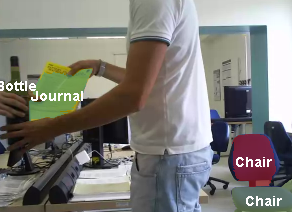}
\includegraphics[width=0.24\textwidth]{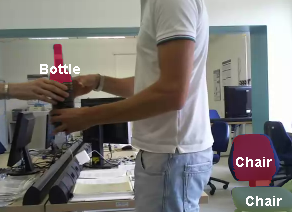}
\caption{Sample predictions and user interactions taken form a real-world video stream. User interactions are marked with red-crosses (in red-framed pictures). The most confident DVA predictions are highlighted only for regions having the highest tag score above $0$.
Only the first two rows refer to the portion of the video during which the user was providing supervisions (70 positive, 2 negative labels).}
\label{fig:labcam}
\end{figure*}

\begin{figure*}
\begin{center}
\includegraphics[width=0.24\textwidth]{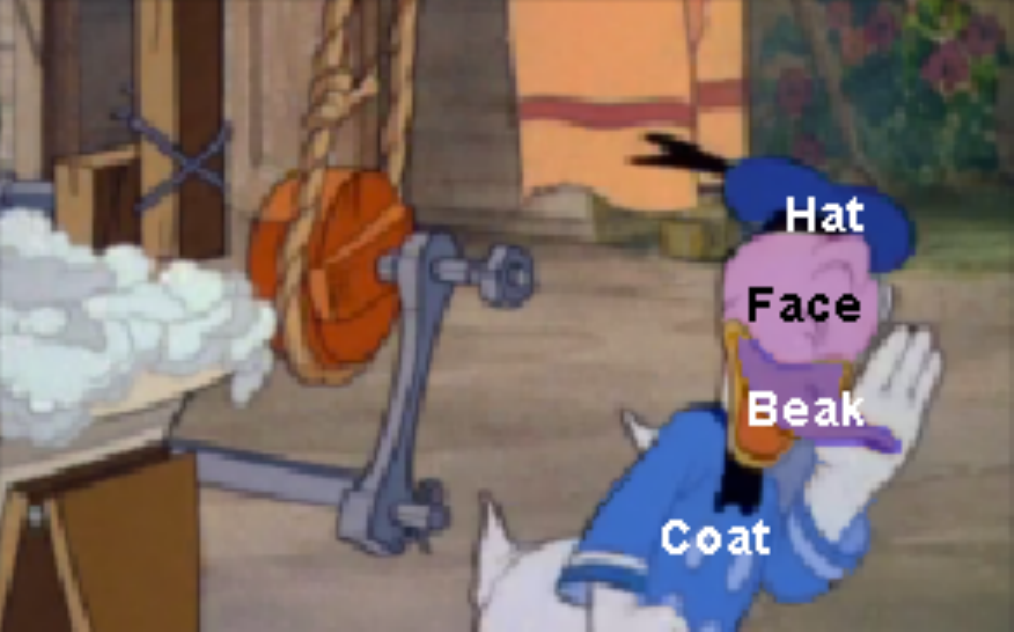}
\includegraphics[width=0.24\textwidth]{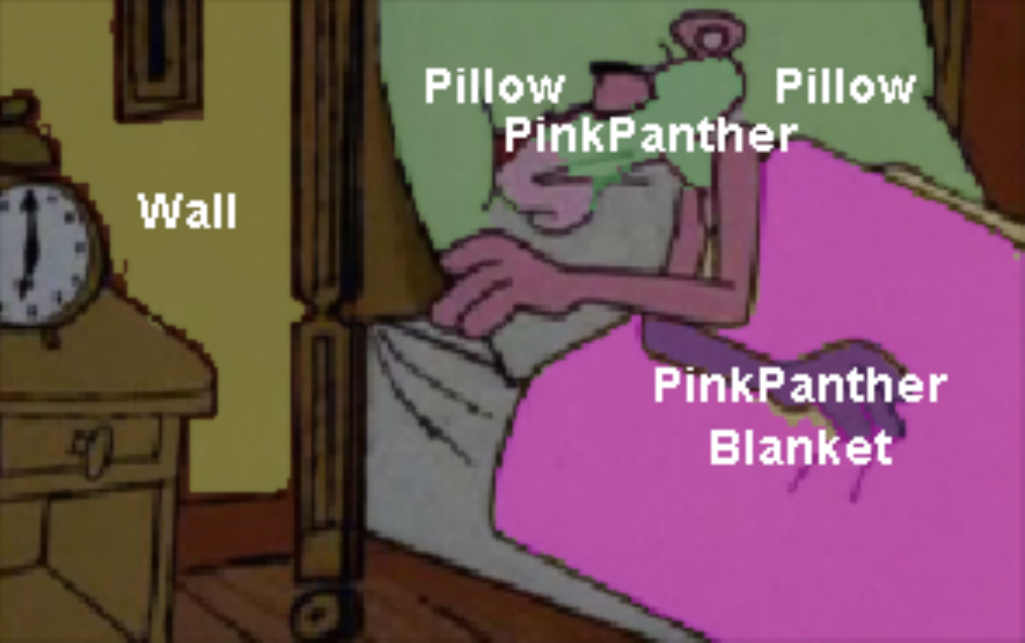}
\includegraphics[width=0.24\textwidth]{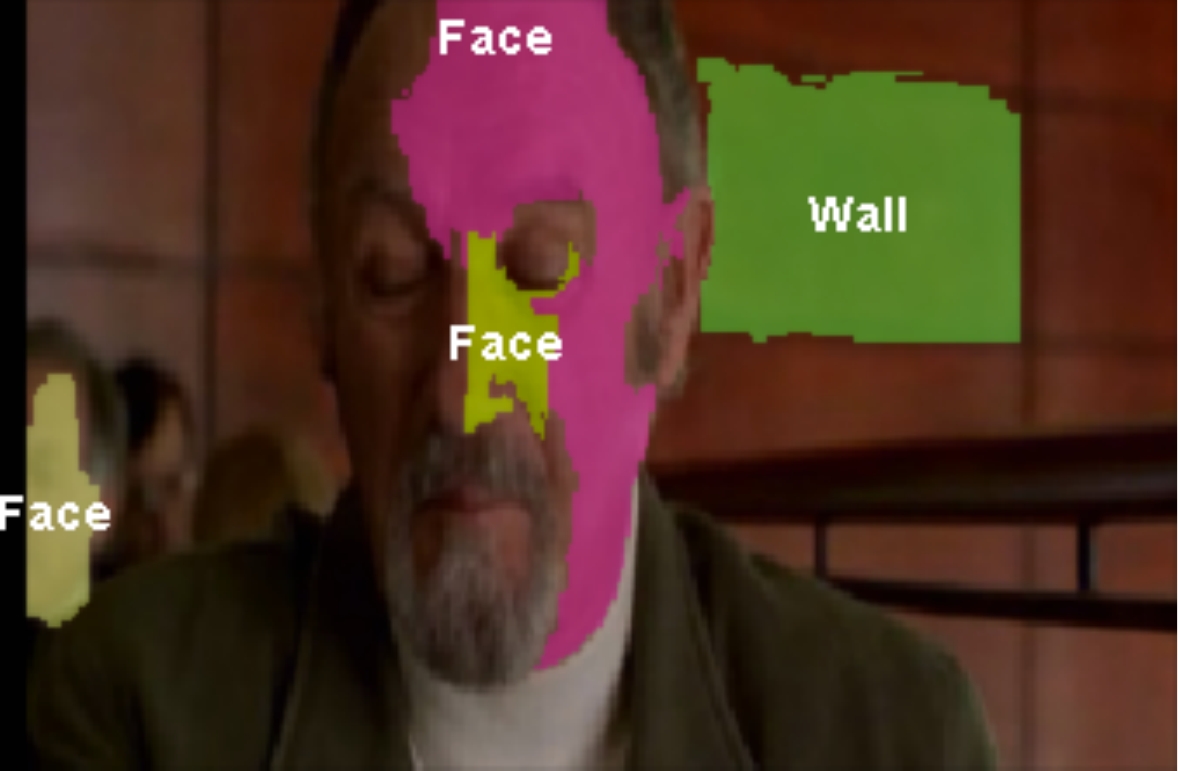}
\includegraphics[width=0.24\textwidth]{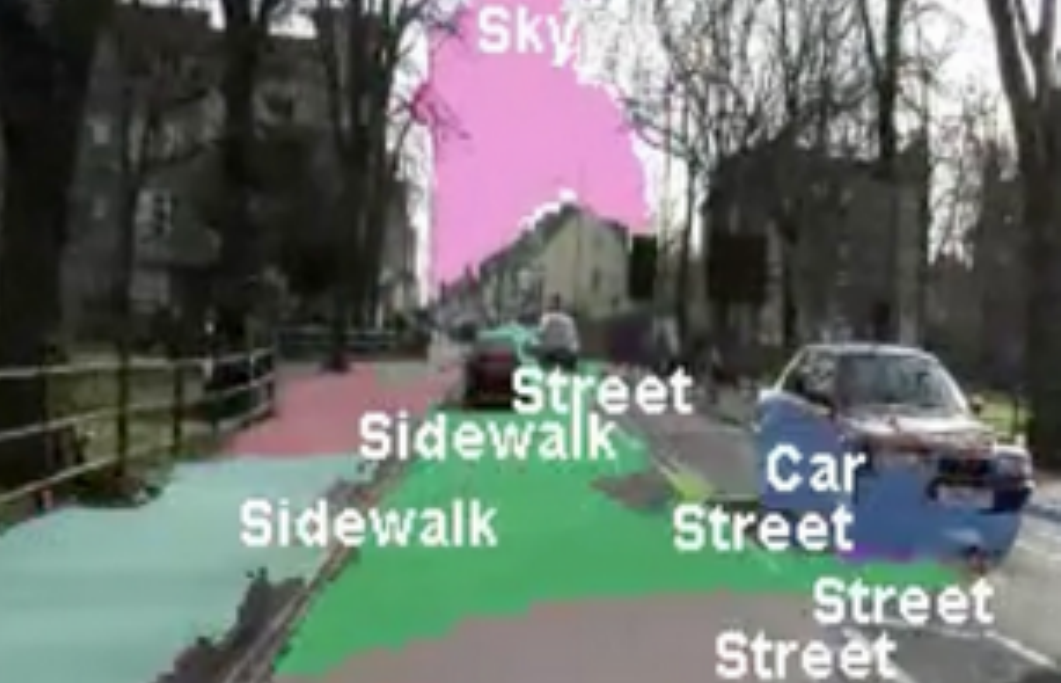}
\caption{Some examples of semantic labeling performed by DVAs on a number of different videos (Donald Duck, \textcopyright\  The Walt Disney Company; Pink Panther, \textcopyright\ Metro Goldwyn Mayer). Only regions where the confidence of the prediction is above a certain threshold are associated with a tag.\label{fig:predictions}}
\end{center}
\end{figure*}

\marginpar{{\em  Artificial \\
and natural \\ visual environments }} 
A visual environment from the AI-lab of UNISI was constructed
using a 2-minute video stream acquired by a webcam. 
During the first portion of the video, a supervisor interacted with the DVA
by providing as many as 72 pixel-wise supervisions, out of which only 2 were negative.
The supervisions covered four object classes (bottle, chair, journal, face). 
In the remaining portion of the video no further supervision is given.
Figure \ref{fig:labcam} collects examples of both user interactions and the most confident DVA predictions, highlighting only regions having the highest tag score above the $0$ threshold. 
The frame sequence is ordered following the real video timeline (left to right, top to bottom). 
The first two rows show samples taken from the first portion of the video, where red-framed pictures mark user supervisions, while the others illustrate DVA's responses. For example, we can observe a wrong ``bottle'' label predicted over the black monitor in the third sample, which is corrected, later on, 
by a subsequent negative supervision.
The last two rows refer to the video portion in which no supervisions were provided. The system is capable of generalizing predictions even in presence of small occlusions (chair, bottle), or in cases where objects appear in contexts that are different from the ones in which they were supervised (bottle, journal, face). 
Humans involved in crowd-sourcing assessment are likely to provide different scores
but it is clear that the learning process of the DVA leads to remarkable performance 
by receiving only a few human supervisions.
Following the same paradigm,  DVAs have been developed on several visual environments, 
ranging from cartoons to movies. Regardless of the visual environment, we use a 
few supervisions, ranging from 1 to 10 per class, for a number of categories between 5 and 10. 
Many of these DVAs can be downloaded at
\url{http://dva.diism.unisi.it} with screenshots and video sequences 
from which the frames of Figure~\ref{fig:predictions} were extracted.

\marginpar{{\em CamVid \\ benchmark}} 
Now, following the standard evaluation scheme, we 
show results on the Cambridge-driving Labeled Video Database (CamVid)~\cite{camvid}. This benchmark consists of a collection of videos captured by a vehicle driving through the city of Cambridge, with ground truth labels associated to each frame from a set of 32 semantic classes. We reproduced the experimental setting employed by almost any work on the CamVid database in recent years~\cite{tighe2013superparsing,lecuncamvid}.
We considered only the 11 most frequent semantic classes, and used only the 600 labeled frames of the dataset (resulting in a video at 1Hz), by splitting them into a training set (367 frames) and a test set (233 frames). For each ground truth region in each frame of the training set, a supervision was provided to DVA, by computing the medoid of the ground truth region and by attaching the supervision to the region constructed by the DVA aggregation process, which contains the medoid pixel. By following this scheme, it is worth mentioning that, with respect to the existing work on the same dataset, DVAs are conceived for online interaction, and not for a massive processing of labeled frames. Therefore, we decided to use a fraction of the available supervisions.
 While almost all the other approaches exploit all the supervised pixels (about 28 millions), 
 we used about 17,000 supervisions, that is more than three order of magnitude less. A variety of different approaches were used on this dataset. The state-of-the-art is obtained by exploiting  Markov Random Fields~\cite{lecuncamvid} or Conditional Random Fields~\cite{tighe2013superparsing}. In this paper, we do not compare
 DVAs against these approaches, because the analysis of a post-processing (or refinement) stage for DVA predictions based on spatial or semantic reasoning is beyond the scope of this work. We therefore compare only against those methods which exploit appearance features, motion and geometric cues.
 We refer to (i)~\cite{cipollaeccv2008}, where bag-of-textons are used as appearance features, and motion and structure properties are estimated from cloud points, and to (ii)~\cite{lecuncamvid}, where several versions of a convolutional neural network (CNN) are tested, either enforcing spatial coherence among superpixels (CNN-superpixels), or weighting the contributions of multilayer predictions with a single scale (CNN-MR fine) or with multiple scales (CNN-multiscale). In order to incorporate the information regarding region positions within the frame, which is an important feature in this scenario, we simply estimated from the training set the a priori probability of each class given the pixel coordinates, and, for each region, we multiplied the score computed by DVA for each class with the prior probability of its centroid. Table~\ref{tab:camvid} shows the results of the experimental comparison. The performance of DVAs are better than CNN-superpixels and are comparable with motion-structures cues, while they are slightly inferior to appearance cues. 
Not surprisingly, given the general  nature of our approach, more specific methods oriented to this task perform better than DVA on most classes. It is the case of the last three competitors in Table~\ref{tab:camvid} that, beside exploiting a larger amount of supervisions, rely on combinations of multiple hypotheses specifically designed for the benchmark. 

\begin{table*}
\begin{center}
\caption{Quantitative evaluation on the CamVid dataset.\label{tab:camvid}}
\vskip 1mm
\footnotesize
\setlength{\tabcolsep}{4pt}
\begin{tabular}{c||c|c|c|c|c|c|c|c|c|c|c||c|c}
Model						& \begin{turn}{90}Building\end{turn} & \begin{turn}{90}Tree\end{turn} & \begin{turn}{90}Sky\end{turn} & \begin{turn}{90}Car\end{turn} & \begin{turn}{90}Sign-Symbol\end{turn} & \begin{turn}{90}Road\end{turn} & \begin{turn}{90}Pedestrian\end{turn} & \begin{turn}{90}Fence\end{turn} & \begin{turn}{90}Column-Pole\end{turn} & \begin{turn}{90}Sidewalk\end{turn} & \begin{turn}{90}Bicyclist\end{turn} & \begin{turn}{90}Average\end{turn} & \begin{turn}{90}Global\end{turn} \\
\hline
DVA						& 28.3 & 34.9 & 95.7 & 31.3 & 22.2 & 91.4 & 54.4 & 29.2 & 11.9 & 74.4 & 13.4 & 44.3 & 63.6 \\
CNN-superpixels~\cite{lecuncamvid}		&  3.2 & 59.7 & 93.5 &  6.6 & 18.1 & 86.5 &  1.9 &  0.8 &  4.0 & 66.0 &  0.0 & 30.9 & 54.8 \\
Motion-Structure cues~\cite{cipollaeccv2008}	& 43.9 & 46.2 & 79.5 & 44.6 & 19.5 & 82.5 & 24.4 & 58.8 &  0.1 & 61.8 & 18.0 & 43.6 & 61.8 \\
Appearance cues~\cite{cipollaeccv2008}		& 38.7 & 60.7 & 90.1 & 71.1 & 51.4 & 88.6 & 54.6 & 40.1 &  1.1 & 55.5 & 23.6 & 52.3 & 66.5 \\
\hline
CNN-MR fine~\cite{lecuncamvid}			& 37.7 & 66.2 & 92.5 & 77.0 & 26.0 & 84.0 & 50.9 & 43.7 & 31.0 & 65.7 & 29.7 & 54.9 & 68.3 \\
CNN-multiscale\cite{lecuncamvid}		& 47.6 & 68.7 & 95.6 & 73.9 & 32.9 & 88.9 & 59.1 & 49.0 & 38.9 & 65.7 & 22.5 & 58.6 & 72.9 \\
Combined cues~\cite{cipollaeccv2008}		& 46.2 & 61.9 & 89.7 & 68.6 & 42.9 & 89.5 & 53.6 & 46.6 &  0.7 & 60.5 & 22.5 & 53.0 & 69.1 \\
\end{tabular}
\end{center}
\end{table*}

\section{Conclusions}
\label{sec:conclusions}
%
In this paper we provide a proof of concept that fully learning-based visual agents can acquire 
visual skills only by living in their own visual environment and by human-like interactions, according to the
``learning to see like children'' communication protocol. 
\marginpar{{\em  L2SLC:   proof \\ of concept}} 
This is achieved by 
DVAs according to a lifelong learning scheme, where the differences between supervised 
and unsupervised learning, and between learning and test sets are dismissed.
The most striking result is that DVAs provide early evidence of capabilities 
of {\em learning in any visual environment by using only a few supervised examples}.
This is mostly achieved by shifting the emphasis on the huge amount of visual information
that becomes available within the L2SLC protocol. Basically, motion coherence 
yields tons of virtual supervisions that are not exploited in most of nowadays state of the art 
approaches.

\marginpar{{\em DVAs:  social \\ evolution}}
The DVAs described in this paper can be improved in different ways, and many issues
are still open. The most remarkable problem that we still need to address by 
appropriate theoretical foundations is the temporal evolution of the agents.
In particularly, the dismissal of the difference between learning and test set, along
with the corresponding classic statistical framework, opens a seemingly unbridgeable gap
with the community of computer vision, which uses to bless scientific contributions on the
basis of appropriate benchmarks on common data bases. However, some recent
influential criticisms on the technical soundness of some benchmarks~\cite{benchmarks}
might open the doors to the crowd-sourcing evaluation proposed in this paper. 
The current version of DVAs can be
extended to perform action recognition, as well as higher level cognitive tasks 
by exploiting logic constraints, that are contemplated in the proposed theoretical framework. 

\marginpar{{\em Machine \\ learning \\ methodologies }}
The need to respond to the  L2SLC protocol has led to a deep network where novel learning
mechanisms have been devised. In particular, we have extended the framework of learning
from constraints to on-line processing of videos. The basic ideas of~\cite{learningfromconstraints}
have been properly adapted to the framework of kernel machines by an on-line
scheme which operates on a transductive environment. 
We are currently pursuing an in-depth reformulation of the theory 
given in~\cite{learningfromconstraints} on the feature manifold driven by the visual data 
as temporal sequences. Instead of using the kernel machine mathematical and algorithmic apparatus, 
in that case the computational model is based on ordinary differential equations on manifolds\footnote{This
seems to be quite a natural solution, that has more solid foundations than  updating schemes
based on kernel methods.}.

\marginpar{{\em En plein air: \\ birth of the \\ movement}} 
No matter what the performance of  DVAs are, this paper 
suggests that other labs can naturally face the challenge of learning to
see like children. This could give rise to the birth of the movement of the 
``en plein air" in computer vision, which could stimulate a paradigm shift
on the way machine learning is used. Hence, the most important contribution of this
paper might not be the specific structure of DVAs, but the computational 
framework in which they are constructed.

  \section*{Acknowledgment}
The results coming from this research couldn't have been achieved without the contribution of many people, who have provided suggestions and supports in different forms. In particular, we thank Salvatore Frandina, Marcello Pelillo, Paolo Frasconi,  Fabio Roli, Yoshua Bengio, Alessandro Mecocci, Oswald Lanz, Samuel Rota Bul\`o, Luciano Serafini, Ivan Donadello, Alberto Del Bimbo, Federico Pernici, and Nicu Sebe. 
\bibliographystyle{plain}
\bibliography{manuscript}

\end{document}